\documentclass[11pt]{article}

\usepackage[margin=1in]{geometry}
\usepackage{amsmath, amssymb, amsthm, bm}
\usepackage{graphicx}
\usepackage{float}
\usepackage{placeins}
\usepackage{booktabs}
\usepackage{caption}
\usepackage{subcaption}
\usepackage{algorithm}
\usepackage{algpseudocode}
\usepackage{multirow}
\usepackage{authblk}
\usepackage[authoryear,round]{natbib}
\usepackage{tikz}
\usetikzlibrary{arrows.meta,positioning}
\usepackage{hyperref}

\hypersetup{
    colorlinks=true,
    linkcolor=blue,
    citecolor=blue,
    urlcolor=blue
}

\title{\textbf{Adapting Neural Posterior Estimators for Simulation-Based Inference in Agent-Based Models}}
\author[1]{Junyao Zhang}
\author[1]{Jinglai Li}
\author[1]{Junqi Tang}
\affil[1]{School of Mathematics, University of Birmingham}
\date{}

\begin{document}

\maketitle

\begin{abstract}

Agent-based models (ABMs) are widely used in economics and social sciences. They represent heterogeneous personal behaviors and their interactions by using simulations, whereas their likelihoods are often intractable for parameter estimations. Therefore, most traditional statistical inference techniques are hindered. Driven by recent advances in probabilistic machine learning, recent works have applied neural networks to achieve likelihood-free inference in large-scale ABMs. Specifically, Neural Posterior Estimation using normalizing flows provide amortized Bayesian inference from simulations, yet a pre-trained estimator may become inaccurate when the target task changes. In this work, we investigate the practicality of adapting a pre-trained conditional normalizing flow through a unified sequential neural posterior estimation test-time training (SNPE-TTT) framework. The framework includes four adaptation strategies: full-parameter adaptation, low-rank adaptation, and two update spaces identified from target-task gradients. We evaluate all four methods across economic simulation models with both tractable and intractable likelihoods. These benchmarks demonstrate that the sequential neural posterior estimation test-time training framework can adapt pre-trained posterior estimators to related inference tasks using a limited budget of target simulations. Posterior recovery and computational benefits depend on the task and adaptation strategy, with lower target-stage cost in some settings but no consistently superior posterior accuracy.

\end{abstract}

\noindent\textbf{Keywords:} Agent-based model; Bayesian inference; Simulation-based inference; Neural posterior estimation; Posterior adaptation; Parameter-efficient adaptation

\section{Introduction}

Agent-based models (ABMs) are broadly applied in studying economic and social science systems. Due to their ability to describe realistic and heterogeneous decisions and interaction rules between agents, these models allow modellers to analyze the whole systems mechanistically with respect to individual \citep{gilbert2019agent,GRAZZINI201726,dyer2024black}. These models represent the macroscopic properties of whole system through stochastic simulations, benefiting modellers to avoid to investigate macro-level phenomena directly, and to study more complex systems. In general, ABMs take model parameters $\theta$ as input, and return time-series simulation data $x_t$, which are drawn from the conditional probability density function $p(x\mid\theta)$. The key of successfully applying ABMs to real-world problems is to perform parameter inference of $\theta$, so that our simulated $x_t$ can be a good match to observed data $y_t$.

Although the stochastic nature of ABMs provides the flexibility, it also makes parameter inference challenging. Even under a fixed parameter setting $\theta$, repeated simulations may produce different outputs $x$, making the likelihood function $p(x\mid\theta)$ analytically intractable or computationally expensive to evaluate. So standard likelihood-based statistical inference approaches, such as maximum likelihood estimation, are unsuitable. Bayesian inference is particularly attractive in this setting because it characterizes uncertainty over the parameter space rather than returning only a point estimate. Consequently, a range of simulation-based inference (SBI), including approximate Bayesian computation and surrogate likelihood methods, have been increasingly applied to economic ABMs. Despite their usefulness, many conventional approaches remain computationally demanding because posterior inference is closely coupled with repeated simulation or likelihood approximation, and some additionally require restrictive assumptions such as stationarity or time-homogeneity of the simulated process \citep{GRAZZINI201726,Lamperti2018surrogates,SHIONO2021104082,platt2022bayesian}. 

Recent advances in machine learning have motivated a new generation of neural simulation-based inference methods that use flexible function approximators to learn the relationship between model parameters and simulated data more directly. Among these approaches, neural posterior estimation (NPE) directly learns a conditional approximation $q_\phi(\theta\mid x)$ to the posterior distribution from simulated parameter-data pairs by using normalizing flows\citep{NIPS2016_6aca9700,cranmer2020frontier}. Its sequential variant, sequential neural posterior estimation (SNPE), further concentrates later simulations on parameter regions that are informative for a particular observation. Once trained, the conditional density estimator can generate posterior samples without repeatedly calling the simulator within a new Markov chain, providing an amortized inference mechanism that is particularly valuable when simulations are computationally expensive. Neural simulation-based inference has consequently shown promising performance for parameter inference in ABMs \citep{wang2024preconditioned}.

However, the same amortization that makes neural posterior estimation computationally efficient can also create a transfer problem. A posterior estimator trained under one simulator configuration learns the relationship between simulated data and parameters under that particular training regime. When the data-generating process at test time deviates from the simulator distribution seen during training, the accuracy of amortized posterior approximations can deteriorate substantially \citep{schmitt2024detecting}. A straightforward solution is to discard the pretrained estimator and retrain a new model from random initialization under the target regime. However, this ignores information already learned from the source simulator and can incur substantial additional simulation and optimization costs.

In this work, we develop a unified SNPE test-time training (SNPE-TTT) framework for parameter inference in ABMs. Starting from a posterior estimator trained under a source regime, SNPE-TTT adapts the estimator to a target inference problem using a limited budget of target simulations. Within the same sequential procedure, we compare several alternative update strategies, including full-parameter adaptation (SNPE-FPA), low-rank adaptation (SNPE-LoRA), and two gradient-informed subspace methods, GradSubspace-FPA and GradSubspace-PEA. We further assess these four posterior adaptation approaches, using distribution-level discrepancy measures that are widely employed in statistics and machine learning.  We compare the adapted posteriors directly against approximate ground truth posteriors in a number of likelihood-tractable models. We further evaluate a likelihood-free Franke--Westerhoff model using simulation-based calibration across independent target observations. Our experiments show that SNPE-TTT can recover key features of target posteriors using a limited budget of target simulations, demonstrating the value of reusing pre-trained estimators. Although no single adaptation strategy dominates across settings and evaluation metrics, the framework supports task-dependent trade-offs among posterior recovery, update-space flexibility and computational cost.

In summary, our contributions with this article are to provide:

\begin{enumerate}
    \item a unified SNPE-TTT framework that reuses pretrained posterior estimators to adapt to related inference tasks under a limited target simulation budget;
    \item a systematic comparison of SNPE-FPA, SNPE-LoRA, GradSubspace-FPA, and GradSubspace-PEA, assessing how different update spaces affect posterior recovery and computational cost;
    \item an evaluation across economic simulation models with tractable and intractable likelihoods, covering simulator-regime and target-observation changes through reference-posterior comparisons and simulation-based calibration, respectively.
\end{enumerate}

\section{Neural Posterior Estimation}

In this section, we provide an overview of Neural Posterior Estimation (NPE), one of the most prominent simulation-based inference method, and its application in parameter estimations for ABMs \citep{dyer2024black, NIPS2016_6aca9700, NIPS2017_addfa9b7, Greenberg20194288}. The core idea underlying NPE is to employ neural networks to approach the parameter posterior density directly via neural conditional density estimators normalizing flow.

\subsection{Normalizing flows}

Normalizing flows have capacity of constructing a flexible density by transforming a simple base distribution into a sophisticated one using neural networks \citep{Tabak2010217, https://doi.org/10.1002/cpa.21423, rezende2015variational,papamakarios2021normalizing}. Given a random variable $\mathbf{U}\sim p_{\mathbf{U}}$ following a simple distribution, that is , we can sample from $p_U$ and evaluate $p_U(u)$ for any $u$ easily, and define
$\mathbf{X}=g(\mathbf{U})$ where $g$ is bijective with inverse $f:=g^{-1}$.
By the change of variables formula, the induced density is
\begin{equation}
p_{\mathbf{X}}(\mathbf{x})
= p_{\mathbf{U}}\!\bigl(f(\mathbf{x})\bigr)\,\left|\det J_f(\mathbf{x})\right|,
\label{eq:nf-cov}
\end{equation}
where $J_f$ denotes the Jacobian of $f$. Sampling from $p_{\mathbf{X}}$ is therefore straightforward by drawing $\mathbf{u}\sim p_{\mathbf{U}}$ and setting $\mathbf{x}=g(\mathbf{u})$. Density evaluation can be obtained from the right-hand side of Equation \eqref{eq:nf-cov} by given $\mathbf{x}$, computing $\mathbf{u}=f(\mathbf{x})$ and the Jacobian determinant term. Practically we can take a simple composition $g = g_n\circ g_{n-1}\circ\cdots\circ g_1,$, to construct a flow, so that one can iteratively transform $\mathbf{u}_0=\mathbf{u}\sim p_{\mathbf{U}}$ via $\mathbf{u}_i=g_i(\mathbf{u}_{i-1})$ to obtain $\mathbf{x}=\mathbf{u}_n$, and evaluate densities by applying the inverses $f_i$ in reverse order while accumulating the corresponding Jacobian determinants, for which we now have Equation \eqref{eq:nf-cov11}

\begin{equation}
q_{\phi}(\mathbf{x})
=
p_{\mathbf{U}}\!\Big(
f_{\phi_1}\big(\cdots f_{\phi_{n-1}}(f_{\phi_n}(\mathbf{x}))\cdots\big)
\Big)
\left|\prod_{i=1}^{n}\det J_{f_{\phi_i}}(\mathbf{x})\right|.
\label{eq:nf-cov11}
\end{equation}

Practically, a standard normal distribution is often token as the simple base distribution, and $f_{\phi}$ are typically neural networks which are designed to be invertible and differentiable. Additionally, for the convenience of Jacobian computation, an architecture of normalising flow that results in a triangular Jacobian has been adopted in previous work \citep{papamakarios2017masked}, so that the determinant is simply the product of the main diagonal elements
Given data samples $\mathbf{x}^{(r)}\stackrel{\mathrm{iid}}{\sim}p_{\mathbf{X}}$, the learnable parameters
$\phi=\{\phi_i:1\le i\le n\}$ of the flows are optimized by maximum log-likelihood:
\begin{equation}
\hat{\phi}
= \arg\max_{\phi}\sum_{r=1}^{R}\log q_{\phi}\!\bigl(\mathbf{x}^{(r)}\bigr).
\label{eq:nf-mle}
\end{equation}

\subsection{Neural posterior estimation with normalizing flows}

Normalizing flows can also be extended to \emph{conditional} density estimation. In ABMs, this enables us to model the posterior directly with a conditional flow$\ q_\phi(\boldsymbol{\theta}\mid \mathbf{x})$\. The trainable parameters $\phi$\ are optimized by maximizing log-likelihood on simulated joint pairs $(\mathbf{x}^{(r)},\boldsymbol{\theta}^{(r)})$ generated from$\ p(\mathbf{x},\boldsymbol{\theta}) = p(\boldsymbol{\theta})\,p(\mathbf{x}\mid \boldsymbol{\theta}),
r=1,\dots,R. $\
Thus, it yields the estimator
\begin{equation}
\hat{\phi}
= \arg\max_{\phi\in\Phi}\sum_{r=1}^{R}\log q_\phi\!\bigl(\boldsymbol{\theta}^{(r)}\mid \mathbf{x}^{(r)}\bigr).
\label{eq:condflow-mle}
\end{equation}
A key advantage of this conditional-flow formulation is that it learns a single amortised posterior estimator across
different $\mathbf{x}$, rather than training a separate density estimator for each fixed observation when evaluating the posterior.

\begin{algorithm}[H]
\caption{Sequential Neural Posterior Estimation}
\label{SNPE}
\begin{algorithmic}[1]

\State \textbf{Input:} Prior distribution $p(\theta)$, simulator $p(x \mid \theta)$, 
observation $y$, conditional density estimator $q_\phi(\theta \mid x)$,
number of rounds $M$, number of simulations per round $N$.

\State Initialize proposal $\tilde{p}_{0}(\theta) = p(\theta)$.
\State Initialize dataset $\mathcal{D} = \{\}$.

\For{$m = 0,\ldots,M-1$}

    \State Sample parameters $\theta_n^{(m)} \sim \tilde{p}_{m}(\theta)$ for $n = 1,\ldots,N$.
    \State Simulate data $x_n^{(m)} \sim p(x \mid \theta_n^{(m)})$.

    \State Append dataset:
    \[
        \mathcal{D} 
        := 
        \mathcal{D} \cup \bigcup_{n=1}^{N} \{ (x_n^{(m)}, \theta_n^{(m)}) \}.
    \]

    \State \textbf{SNPE training step}
    \While{not converged}

        \State Evaluate loss:
        \[
        \mathcal{L}(\phi) 
        = 
        - \sum_{(x,\theta)\in\mathcal{D}}
            \log \Bigg(
                q_\phi(\theta \mid x)
                \frac{ \tilde{p}_{m}(\theta) }{ p(\theta) }
                \frac{1}{Z(x, \phi)}
            \Bigg).
        \]

        \State Update trainable parameters $\phi$ using $\nabla_\phi \mathcal{L}(\phi)$.

    \EndWhile

    \State Set new proposal distribution:
    \[
        \tilde{p}_{m+1}(\theta)
        := 
        q_\phi(\theta \mid y).
    \]

\EndFor
\end{algorithmic}
\end{algorithm}
\subsection{Sequential neural posterior estimation}

In NPE, most parameters drawn from single prior make simulations far from the posterior support for a fixed observation $\mathbf{x}$, leading to wasted simulation budget. Sequential neural posterior estimation (SNPE) instead uses multiple rounds to focus simulations on parameter regions relevant to that observation \citep{Greenberg20194288,lueckmann2021benchmarking}.
Let $\tilde p_m(\boldsymbol{\theta})$ denote the proposal used at round $m$, with initialization
$\tilde p_0(\boldsymbol{\theta})=p(\boldsymbol{\theta})$.
At each round, we simulate
\[
\boldsymbol{\theta}^{(n)} \stackrel{\mathrm{iid}}{\sim} \tilde p_m(\boldsymbol{\theta}),
\qquad
\mathbf{x}^{(n)} \sim p(\mathbf{x}\mid \boldsymbol{\theta}^{(n)}),
\]
append the resulting pairs to the training set, and then retrain learnable network parameters. A common choice is to set the next proposal to the current posterior estimate at the target observation,
\[
\tilde p_{m+1}(\boldsymbol{\theta}) := q_\phi(\boldsymbol{\theta}\mid \mathbf{y}),
\]
so that successive rounds concentrate simulations around high-posterior-mass regions.

From round $m\ge 1$, the training data are no longer drawn from the prior joint $p(\mathbf{x},\boldsymbol{\theta})$. The objective must therefore correct for proposal sampling. For the posterior-transformation formulation used here, the loss is
\begin{equation}
\mathcal{L}(\phi)
= - \sum_{(\mathbf{x},\boldsymbol{\theta})\in\mathcal{D}}
\log\!\left(
q_\phi(\boldsymbol{\theta}\mid\mathbf{x})
\frac{\tilde p_m(\boldsymbol{\theta})}{p(\boldsymbol{\theta})}
\frac{1}{Z(\mathbf{x},\phi)}
\right),
\label{eq:snpe-round-objective}
\end{equation}
where $Z(\mathbf{x},\phi)$ normalizes the transformed density over $\boldsymbol{\theta}$. Algorithm~\ref{SNPE} summarizes the version used in the experiments.

\section{Test-Time training Framework for Sequential Neural Posterior Estimation }
\label{sec:ttt_framework}

Training an amortized posterior estimator with SNPE can require substantial simulation and computational resources. This motivates us to consider reusing a pre-trained estimator for a related similar target inference task with a limited simulation budget. Let $q_{\phi_0}(\theta\mid x)$ denote the original estimator, with parameters $\phi_0$. Given a target simulator $p_1(x\mid\theta)$, a prior $p(\theta)$ and an observation $y$, the objective is to approximate
\begin{equation}
    p_1(\theta\mid y)
    \propto p(\theta)p_1(y\mid\theta).
    \label{eq:target_posterior}
\end{equation}
The original and target tasks share the parameter and observation representations required by the posterior network.

Therefore, we proposed test-time training framework within SNPE (SNPE-TTT), which initializes target adaptation from $\phi_0$ and sequentially acquires target simulations. Here, test-time training refers to simulator-supervised adaptation after original training and requires continued access to the target simulator. At target round $m$, parameters are drawn from a proposal $\tilde p_m(\theta)$ and passed to the target simulator. The resulting pairs are added to the adaptation dataset, and the estimator is updated using a proposal-corrected SNPE objective. The next proposal is the updated posterior conditioned on $y$. The first proposal is the prior, and the total target simulation budget is $\sum_{m=0}^{M-1}N_m$.

Figure~\ref{fig:snpe_ttt_framework} summarizes this procedure. We consider four adaptation strategies: SNPE-FPA, SNPE-LoRA, GradSubspace-FPA and GradSubspace-PEA. They specify which parameters or coordinates are optimized and, where applicable, how the adaptation basis changes across rounds.

\begin{figure}[!htbp]
\centering
\resizebox{\linewidth}{!}{%
\begin{tikzpicture}[
    node distance=7mm and 7mm,
    box/.style={
        draw, rounded corners, align=center,
        minimum height=9mm, text width=29mm,
        fill=blue!5
    },
    choice/.style={
        draw, rounded corners, align=center,
        minimum height=12mm, text width=43mm,
        fill=orange!10
    },
    arrow/.style={-{Latex[length=2mm]}, thick}
]
\node[box] (source) {Original simulator\\and simulations};
\node[box, right=of source] (pretrain)
    {Original posterior\\estimator $q_{\phi_0}$};
\node[box, below=of source] (target)
    {Target simulator\\and observation $y$};
\node[box, right=of target] (pairs)
    {Sequential target\\parameter--data pairs};
\node[choice, right=of pairs] (space)
    {Adaptation strategy\\
     SNPE-FPA $\mid$ SNPE-LoRA\\
     GradSubspace-FPA $\mid$ GradSubspace-PEA};
\node[box, right=of space] (adapt)
    {Optimize target\\SNPE objective};
\node[box, above=of adapt] (posterior)
    {Adapted posterior\\$q_{\phi^\star}(\theta\mid y)$};

\draw[arrow] (source) -- (pretrain);
\draw[arrow] (target) -- (pairs);
\draw[arrow] (pretrain) -| (space);
\draw[arrow] (pairs) -- (space);
\draw[arrow] (space) -- (adapt);
\draw[arrow] (adapt) -- (posterior);
\draw[arrow] (adapt.south)
    |- ++(0,-7mm) -| (pairs.south)
    node[pos=0.55, below] {next-round proposal};
\end{tikzpicture}%
}
\caption{The SNPE-TTT framework. A source-trained posterior estimator is adapted using sequentially acquired target simulations. The selected strategy determines the trainable parameters, update constraints and basis-refresh schedule.}
\label{fig:snpe_ttt_framework}
\end{figure}

\subsection{Full-parameter and low-rank adaptation}

\subsubsection{Full-parameter adaptation}

SNPE with full-parameter adaptation (SNPE-FPA) updates all trainable parameters of the original posterior estimator using the target SNPE objective without additional constraint on the parameter update space. Adam maintains first- and second-moment states for every trainable parameter.

\subsubsection{Low-rank adaptation}

SNPE with low-rank adaptation (SNPE-LoRA) freezes the previous weights and learns additive low-rank updates \citep{hu2022lora}. For a weight matrix $W_0\in\mathbb R^{d\times k}$, the adapted weight is
\begin{equation}
    W'=W_0+sBA,
    \qquad
    A\in\mathbb R^{r\times k},
    \quad
    B\in\mathbb R^{d\times r},
    \label{eq:lora_update}
\end{equation}
where $r$ is the adapter rank and $s=\alpha/r$ is a fixed scaling factor. The experiments use $\alpha=r$, giving $s=1$. We initialize the factors as
\begin{equation}
    A_{uv}\sim\mathcal N(0,\sigma_A^2),
    \qquad B=0.
\end{equation}
The initial update is therefore zero, preserving the original estimator at the beginning of adaptation. This parameterization constrains the rank of each weight update, while its row and column spaces can change as both factors are learned. It does not restrict adaptation to a predetermined linear subspace.

\subsection{Gradient-informed subspace adaptation}
\label{sec:gradsubspace}

We also construct adaptation bases explicitly from gradients of the target SNPE objective, building on gradient-subspace fine-tuning methods \citep{duanlifelong,zhang2023finetuninghappenstinysubspaces,rajabi2024optimizing}. These bases define reduced coordinates for optimization. GradSubspace-FPA refreshes the bases each round and applies lifted updates to the network weights. GradSubspace-PEA retains the first-round bases and optimizes coordinate matrices around the fixed source weights.

\subsubsection{Blockwise subspace construction}

Let $\mathcal J$ index eligible two-dimensional weight tensors, with $W_j\in\mathbb R^{m_j\times n_j}$. Biases and one-dimensional parameters are frozen in both gradient-subspace strategies. For a requested rank $r$, define
\begin{equation}
    \bar r_j=\min(r,m_j,n_j).
\end{equation}
A block is retained only when
\begin{equation}
    \bar r_j(m_j+n_j)<m_jn_j.
\end{equation}
This criterion compares the size of a basis-and-coordinate representation with the corresponding weight matrix; it does not establish a reduction in total model storage.

For each retained block, we collect $K$ mini-batch gradients of the target SNPE loss,
\begin{equation}
    G_{j,b}
    =
    \nabla_{W_j}\mathcal L_m(\phi;\mathcal B_b),
    \qquad b=1,\ldots,K.
    \label{eq:grad_snap}
\end{equation}
GradSubspace-FPA constructs these bases at the start of each round using the current estimator. GradSubspace-PEA constructs them only in the first round, before coordinate adaptation. The loss used for basis construction is the same SNPE loss used for training.

The gradients are not mean-centred, so the construction retains information about their mean direction. If $m_j\ge n_j$, we form
\begin{equation}
    C_j^L
    =
    \frac{1}{K}
    \sum_{b=1}^{K}G_{j,b}G_{j,b}^{\top}
    \label{eq:left_grad_moment}
\end{equation}
and select its leading orthonormal eigenvectors. Otherwise, we form
\begin{equation}
    C_j^R
    =
    \frac{1}{K}
    \sum_{b=1}^{K}G_{j,b}^{\top}G_{j,b}
    \label{eq:right_grad_moment}
\end{equation}
and select the corresponding right basis.

Numerically zero eigen-directions are discarded. Let $\rho_j\le\bar r_j$ denote the resulting effective rank, giving
$U_j\in\mathbb R^{m_j\times\rho_j}$ or
$V_j\in\mathbb R^{n_j\times\rho_j}$.
The construction is blockwise, rather than a single subspace of the flattened network parameters.

For the left-sided case, the selected basis solves the projection problem
\begin{equation}
    U_j\in
    \underset{U^\top U=I_{\rho_j}}{\arg\min}
    \sum_{b=1}^{K}
    \left\|G_{j,b}-UU^\top G_{j,b}\right\|_F^2.
    \label{eq:gradient_projection_objective}
\end{equation}
The right-sided construction has the analogous interpretation. Thus, the basis preserves sampled local gradient energy at the selected dimension. This property alone does not guarantee accurate posterior recovery.

\subsubsection{GradSubspace-FPA: round-wise basis refresh}

GradSubspace-FPA compresses each gradient, applies Adam in reduced coordinates, and maps the resulting update back to the weight matrix. This follows the reduced-optimizer-state principle of gradient low-rank projection methods \citep{zhao2024galorememoryefficientllmtraining}.

For $m_j\ge n_j$, the update is
\begin{equation}
    H_{j,t}=U_j^\top G_{j,t},
    \qquad
    W_{j,t+1}
    =
    W_{j,t}-\eta U_j\mathcal A_t(H_{j,t}),
    \label{eq:proj_grad}
\end{equation}
where $\mathcal A_t$ denotes the bias-corrected Adam transformation. Its moment states have shape $\rho_j\times n_j$. For $m_j<n_j$,
\begin{equation}
    H_{j,t}=G_{j,t}V_j,
    \qquad
    W_{j,t+1}
    =
    W_{j,t}-\eta\mathcal A_t(H_{j,t})V_j^\top,
    \label{eq:update_ttt}
\end{equation}
with moment states of shape $m_j\times\rho_j$.

Applying Adam before lifting ensures that the final update lies in the selected matrix subspace. Each basis remains fixed within a round and is reconstructed at the beginning of the next round. Updated network weights carry over between rounds. The FPA designation refers to applying lifted updates to the stored weight matrices; it does not imply the unrestricted updates used by SNPE-FPA.

\subsubsection{GradSubspace-PEA: fixed-basis coordinate adaptation}

GradSubspace-PEA retains the bases estimated from first-round target simulations and freezes the source weights throughout adaptation. Each eligible matrix is parameterized as
\begin{equation}
    W_j(A_j)=
    \begin{cases}
        W_{j,0}+U_jA_j,
        & m_j\ge n_j,\quad
          A_j\in\mathbb R^{\rho_j\times n_j},\\
        W_{j,0}+A_jV_j^\top,
        & m_j<n_j,\quad
          A_j\in\mathbb R^{m_j\times\rho_j}.
    \end{cases}
    \label{eq:param_peft}
\end{equation}
The coordinate matrices are initialized at zero and are the only trainable parameters. Their total dimension is
\begin{equation}
    d_{\mathrm{PEA}}
    =
    \sum_{\substack{j\in\mathcal J\\m_j\ge n_j}}
        \rho_jn_j
    +
    \sum_{\substack{j\in\mathcal J\\m_j<n_j}}
        m_j\rho_j.
    \label{eq:obj_peft}
\end{equation}
Coordinate values carry over between rounds, while the bases remain fixed. This reduces trainable dimensionality and optimizer state, although storage also includes the source weights and fixed bases.

For a fixed basis, the lifted update and coordinate parameterization are algebraically equivalent under identical initial models, mini-batches and coordinate optimizer states. For example, the chain rule gives
\begin{equation}
    \nabla_{A_j}\mathcal L(W_{j,0}+U_jA_j)
    =
    U_j^\top\nabla_{W_j}\mathcal L.
\end{equation}
This equivalence also requires consistent treatment of gradient clipping and any regularization. In the implemented training protocol, all methods reset Adam and its moment states at each round. The defining distinction is therefore basis refresh: GradSubspace-FPA can accumulate updates across successive subspaces, whereas GradSubspace-PEA remains in one source-anchored affine subspace.

\subsection{Unified adaptation algorithm}

Algorithm~\ref{alg:snpe_ttt_framework} combines the shared sequential procedure with the four adaptation strategies. The target loss $\mathcal L_m$ follows the posterior-transformation principle in Eq.~\eqref{eq:snpe-round-objective}. All strategies start from the same SNPE estimators and follow the same target simulation schedule. Appendix~\ref{app:snpe} and ~\ref{app:experiments} will provide more details.

\begin{algorithm}[htbp]
\caption{Unified SNPE-TTT framework}
\label{alg:snpe_ttt_framework}
\begin{algorithmic}[1]
\State \textbf{Input:} prior $p(\theta)$; target simulator
$p_1(x\mid\theta)$; observation $y$; source parameters $\phi_0$;
strategy $a\in\{\textsc{SNPE-FPA},\textsc{SNPE-LoRA},
\textsc{GradSubspace-FPA},\textsc{GradSubspace-PEA}\}$;
round budgets $N_0,\ldots,N_{M-1}$.
\State Set $\tilde p_0(\theta)=p(\theta)$,
$\mathcal D=\emptyset$, and $\phi\leftarrow\phi_0$.
\If{$a=\textsc{SNPE-LoRA}$}
    \State Freeze source parameters and initialize adapters with
    random $A$ and $B=0$.
\EndIf
\For{$m=0,\ldots,M-1$}
    \State Draw $\theta_n^{(m)}\sim\tilde p_m(\theta)$ and
    $x_n^{(m)}\sim p_1(x\mid\theta_n^{(m)})$
    for $n=1,\ldots,N_m$.
    \State Append the simulated pairs to $\mathcal D$, retaining
    the information required for proposal correction.
    \State Construct the target SNPE loss $\mathcal L_m$.
    \If{$a=\textsc{GradSubspace-FPA}$}
        \State Estimate round-specific blockwise bases from
        $K$ target-loss gradient mini-batches.
    \ElsIf{$a=\textsc{GradSubspace-PEA}$ and $m=0$}
        \State Estimate fixed blockwise bases and initialize
        coordinate matrices $A_j=0$.
    \EndIf
    \State Initialize Adam with zero moment states for the
    parameters or coordinates optimized in this round.
    \If{$a=\textsc{SNPE-FPA}$}
        \State Optimize all learnable network parameters.
    \ElsIf{$a=\textsc{SNPE-LoRA}$}
        \State Optimize only the adapter factors.
    \ElsIf{$a=\textsc{GradSubspace-FPA}$}
        \State Apply Adam to compressed gradients and lift
        the updates to the eligible weight matrices.
    \Else
        \State Optimize only the coordinate matrices,
        keeping source weights and bases fixed.
    \EndIf
    \State Set
    $\tilde p_{m+1}(\theta)\leftarrow q_\phi(\theta\mid y)$.
\EndFor
\State \textbf{Output:} adapted posterior
$q_{\phi^\star}(\theta\mid y)$.
\end{algorithmic}
\end{algorithm}

\section{Numerical Experiments for tractable examples}
\label{sec:experiments}

In this section, we evaluate the performance of four adaptation SNPE-TTT approaches against approximate ground truth posterior sampled by MH, within two likelihood-tractable ABMs under distribution shift. Network and SNPE training details are also provided in Appendix~\ref{app:snpe}, additional experimental details and results in Appendix~\ref{app:experiments}.

\subsection{Performance Metrics}

Each of the following examples in this section has a tractable likelihood function, which enables us to obtain an approximate ground-truth posterior via standard MCMC. To quantify the discrepancy between the estimated posterior \(\hat{p}(\theta \mid x)\) and the ground-truth posterior \(p^\star(\theta \mid x)\), we adopt two probability metrics.

\paragraph{Wasserstein distance.}
This distance measure was derived from optimal transport theory \citep{d2eb0123-2ebc-3952-b609-177897a44f40}, and then broadly used in various machine learning study, such as generative adversarial networks \citep{peyre2019transport}. For two sets of posterior samples
$\{\theta_i\}_{i=1}^n \overset{\mathrm{iid}}{\sim} P$
and
$\{\theta'_j\}_{j=1}^m \overset{\mathrm{iid}}{\sim} Q$,
the Wasserstein metric is defined as
\begin{equation}
\mathcal{W}_p
\left(
\{\theta_i\}_{i=1}^n,
\{\theta'_j\}_{j=1}^m
\right)^p
=
\inf_{\gamma \in \Gamma_{n,m}}
\sum_{i=1}^n
\sum_{j=1}^m
\rho_0(\theta_i,\theta'_j)^p
\gamma_{ij},
\label{eq:wass_definition}
\end{equation}
where $\Gamma_{n,m}$ is the set of non-negative $n\times m$
matrices with columns summing to $m^{-1}$ and rows summing to
$n^{-1}$. We use the Euclidean distance for $\rho_0$ and refer to this metric as WASS. A smaller WASS indicates stronger agreement between the estimated and ground-truth posterior distributions.

\paragraph{Maximum Mean Discrepancy (MMD).}
Let $P$ and $Q$ denote the ground-truth and estimated posterior distributions, respectively, and denote $k$ as a symmetric positive semi-definite kernel. The MMD between $P$ and $Q$ is
\begin{equation}
\mathrm{MMD}(P,Q)
=
\left\|
\mathbb{E}_{\theta\sim P}[k(\theta,\cdot)]
-
\mathbb{E}_{\theta'\sim Q}[k(\theta',\cdot)]
\right\|_{\mathcal H},
\label{eq:mmd_definition}
\end{equation}
where $\mathcal H$ is the reproducing kernel Hilbert space associated
with $k$ \citep{JMLR:v13:gretton12a}. Given posterior samples
$\{\theta_i\}_{i=1}^n \overset{\mathrm{iid}}{\sim} P$
and
$\{\theta'_j\}_{j=1}^m \overset{\mathrm{iid}}{\sim} Q$,
we compute the unbiased empirical estimator
\begin{equation}
\widehat{\mathrm{MMD}}(P,Q)
=
\frac{1}{m(m-1)}
\sum_{\substack{i,j\\j\neq i}}
k(\theta'_i,\theta'_j)
+
\frac{1}{n(n-1)}
\sum_{\substack{i,j\\i\neq j}}
k(\theta_i,\theta_j)
-
\frac{2}{nm}
\sum_{i,j}
k(\theta_i,\theta'_j).
\label{eq:mmd_empirical}
\end{equation}

We use a Gaussian kernel
\begin{equation}
k(\theta,\theta')
=
\exp\left(
-\frac{\|\theta-\theta'\|_2^2}{2\sigma^2}
\right),
\label{eq:mmd_gaussian_kernel}
\end{equation}
where $\sigma^2 = \operatorname{median}\{\|\tilde{\theta}_i -
\tilde{\theta}_j\|_2^2\}$, with
$\{\tilde{\theta}_i\}_{i=1}^n$ denoting samples from the
ground-truth posterior \citep{briol2019statistical}.
Lower MMD indicates closer agreement between the two posterior distributions.

Both metrics are computed between samples from the estimated posterior
$\hat{p}(\theta\mid x)$ and samples from the approximate ground-truth
posterior $p^\star(\theta\mid x)$ obtained using MH.

\subsection{Brock--Hommes Model}
\label{sec:bh-design}

For our first numerical experiment, we use a variant of the Brock-Hommes model \citep{brock1998heterogeneous}, which has been widely consider as ABM calibration studies \citep{platt2020comparison,Lamperti2018surrogates,dyer2024black}. The model describes asset-price dynamics generated by heterogeneous traders who adapt between forecasting rules. The variable $x_t$ denotes the deviation of the asset price from its fundamental value. The fraction of type $h$ traders at time $t+1$ is determined by a discrete-choice rule based on past performance:
\begin{equation}
\begin{aligned}
x_{t+1}
&=
\frac{1}{R}
\left[
\sum_{h=1}^{H}
n_{h,t+1}\bigl(g_h x_t + b_h\bigr)
+
\varepsilon_{t+1}
\right],
\qquad
\varepsilon_{t+1}\sim\mathcal{N}(0,\sigma^2),
\\[6pt]
n_{h,t+1}
&=
\frac{
\exp\!\left[
\beta (x_t - R x_{t-1})(g_h x_{t-2} + b_h - R x_{t-1})
\right]
}{
\sum_{k=1}^{H}
\exp\!\left[
\beta (x_t - R x_{t-1})(g_k x_{t-2} + b_k - R x_{t-1})
\right]
}.
\end{aligned}
\label{eq:bh_dynamics}
\end{equation}

Here, $R=1+r$ is the gross interest rate, $g_h$ is the trend coefficient, $b_h$ is the forecasting bias, and $\beta>0$ is the intensity of choice. We set $H=4$ and fix \(g_1=b_1=b_4=0, g_4=1.01.\) The inferred parameter vector and its prior are \(\theta=(g_2, b_2, g_3, b_3), g_2, b_2, g_3\sim\mathcal U(0,1), b_3\sim\mathcal U(-1,0).\)

Rewriting the Brock--Hommes dynamics in Equations~\ref{eq:bh_dynamics}, we can obtain the transition density
\begin{equation}
\label{eq:bh_transition_f}
\begin{aligned}
p(y_{t+1}\mid y_{1:t},\theta)
&=
\mathcal{N}\!\left(
y_{t+1};
f(y_{t-2:t};\theta),
\frac{\sigma^2}{R^2}
\right),
\\[6pt]
f(y_{t-2:t};\theta)
&=
\frac{1}{R}\sum_{h=1}^{H}
\frac{
\exp\!\left\{
\beta(y_t-Ry_{t-1})(g_hy_{t-2}+b_h-Ry_{t-1})
\right\}
}{
\sum_{h'=1}^{H}
\exp\!\left\{
\beta(y_t-Ry_{t-1})(g_{h'}y_{t-2}+b_{h'}-Ry_{t-1})
\right\}
}
(g_hy_t+b_h).
\end{aligned}
\end{equation}
Thus, we can evaluate this likelihood numerically and use MH to obtain an approximate ground-truth posterior as reference for later analysis, which would be further discussed in Appendix~\ref{app:references}.

\subsubsection{Model Regime shift}

\begin{figure}[!htbp]
\centering
\begin{subfigure}{0.45\linewidth}
\centering
\includegraphics[width=\linewidth]{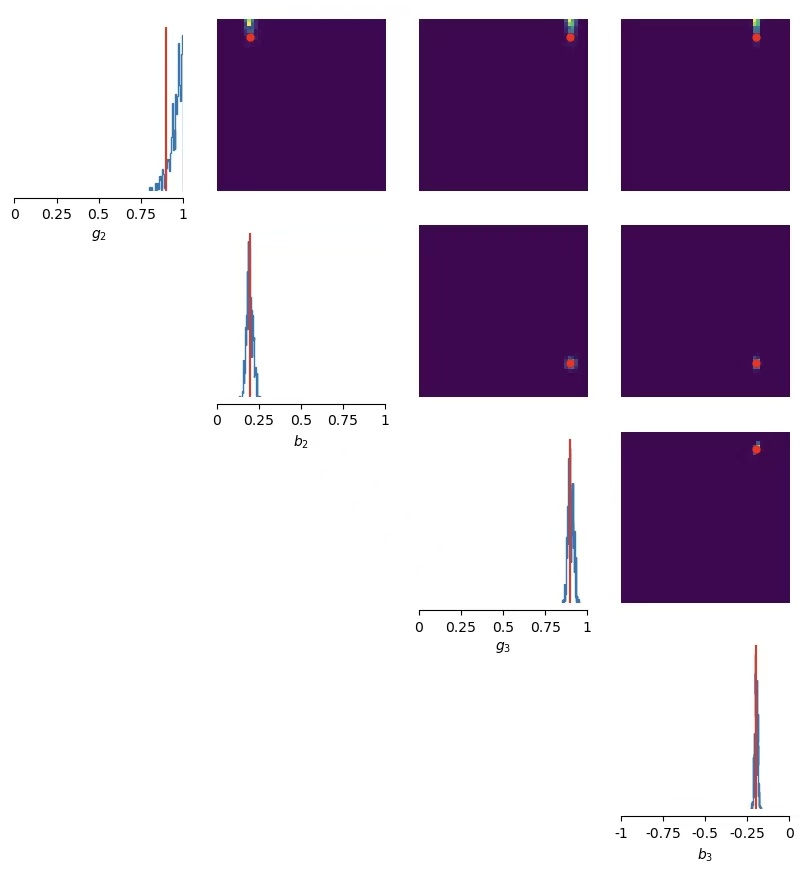}
\caption{Source model ($\beta=120$)}
\end{subfigure}
\hfill
\begin{subfigure}{0.45\linewidth}
\centering
\includegraphics[width=\linewidth]{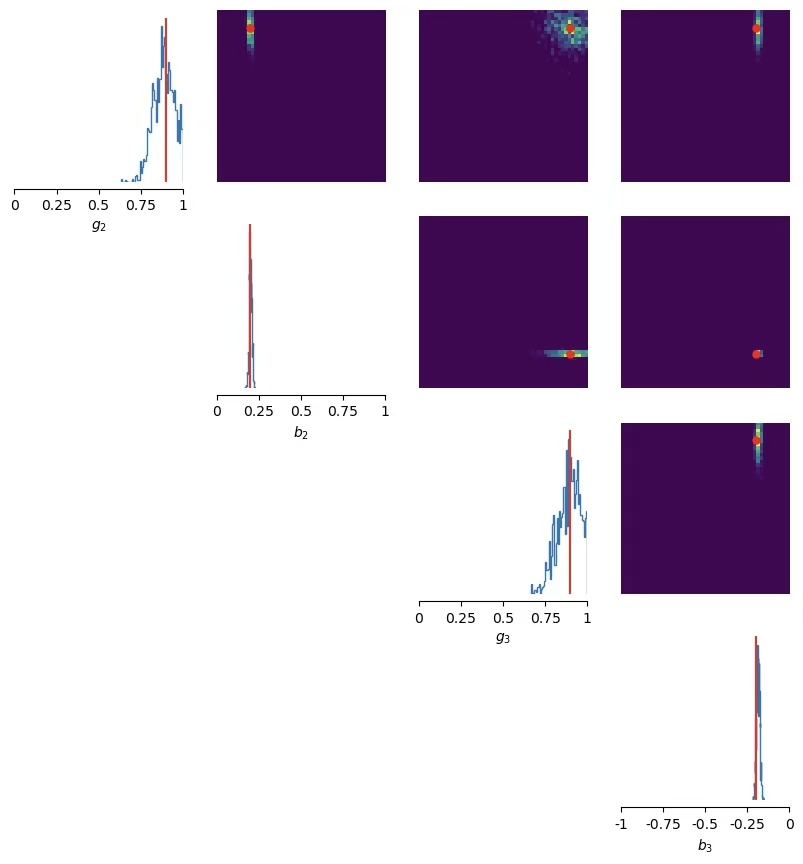}
\caption{Target model ($\beta=60$)}
\end{subfigure}
\caption{Approximate ground-truth posteriors for the Brock-Hommes source and target models.}
\label{fig:pairplotsgt}
\end{figure}

The initial estimator is trained at $\beta_0=120$ and adapted to a target simulator with $\beta_1=60$. Both observations are generated at the fixed structural parameter
\(
\theta^\star=(g_2,b_2,g_3,b_3)=(0.9,0.2,0.9,-0.2),
\)
which means this experiment isolates a change in the intensity of choice rather than a change in the data-generating parameter. The intensity of choice $\beta$ controls how sensitively agents switch between forecasting strategies and is central to empirical heuristic-switching models \citep{kukacka2023heuristic}. Writing
\begin{equation}
n_{h,t}(\beta,\theta)
=
\frac{\exp(\beta U_{h,t}(\theta))}
{\sum_k\exp(\beta U_{k,t}(\theta))},
\end{equation}
where
\(
U_{h,t}(\theta)
=
(x_t-Rx_{t-1})(g_hx_{t-2}+b_h-Rx_{t-1}),
\)
makes the role of $\beta$ explicit. As $\beta\to0$, $n_{h,t}\to1/H$ and strategy choice becomes less sensitive to performance differences. As $\beta$ becomes large, the rule approaches discrete selection of the best-performing strategy. Since two models likelihoods
\(
L_{\beta_0}(y\mid\theta)
\text{and}\ L_{\beta_1}(y\mid\theta)
\)
can therefore have different local geometry as shown in Figure~\ref{fig:pairplotsgt}. 

In general,
\[
\nabla_\theta\log L_{\beta_0}(y_{\beta_1}\mid\theta)
\neq
\nabla_\theta\log L_{\beta_1}(y_{\beta_1}\mid\theta).
\]
Changing $\beta$ can consequently alter posterior shape, so a SNPE trained at $\beta=120$ need not remain calibrated at $\beta=60$. 

\begin{figure}[!htbp]
\centering
\begin{subfigure}{0.45\linewidth}
\centering
\includegraphics[width=\linewidth]{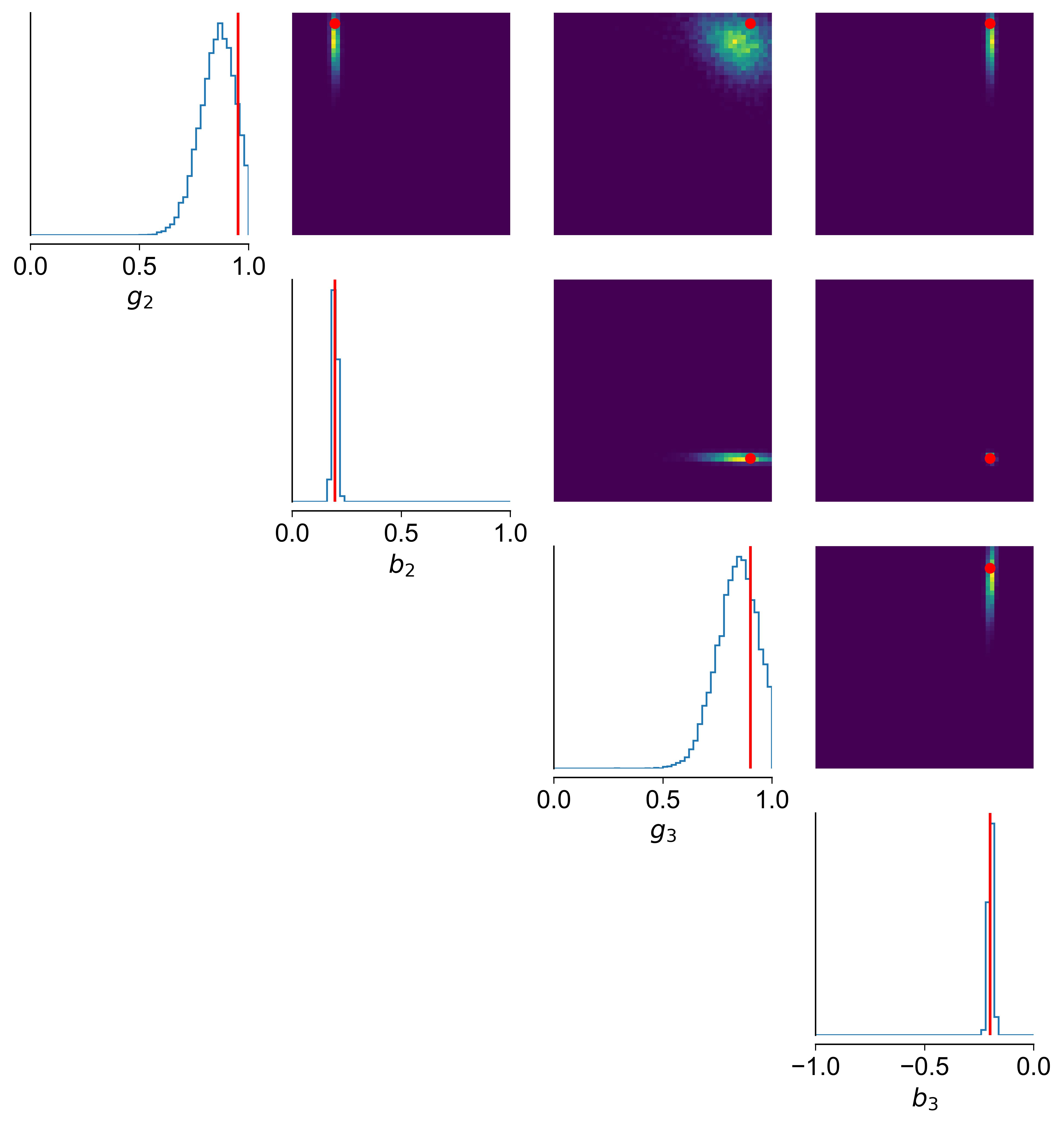}
\caption{SNPE-FPA}
\end{subfigure}
\hfill
\begin{subfigure}{0.45\linewidth}
\centering
\includegraphics[width=\linewidth]{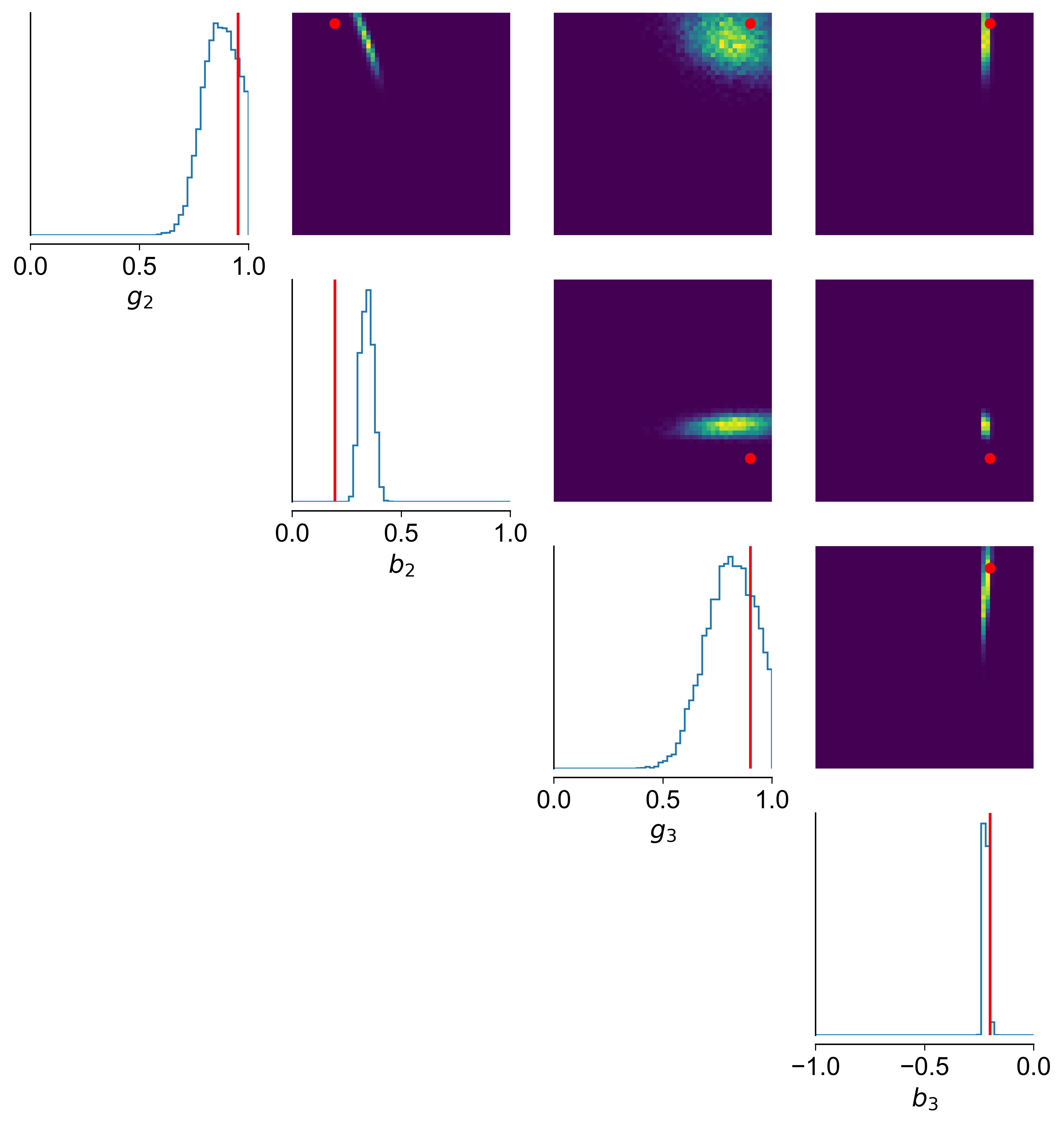}
\caption{SNPE-LoRA}
\end{subfigure}
\begin{subfigure}{0.45\linewidth}
\centering
\includegraphics[width=\linewidth]{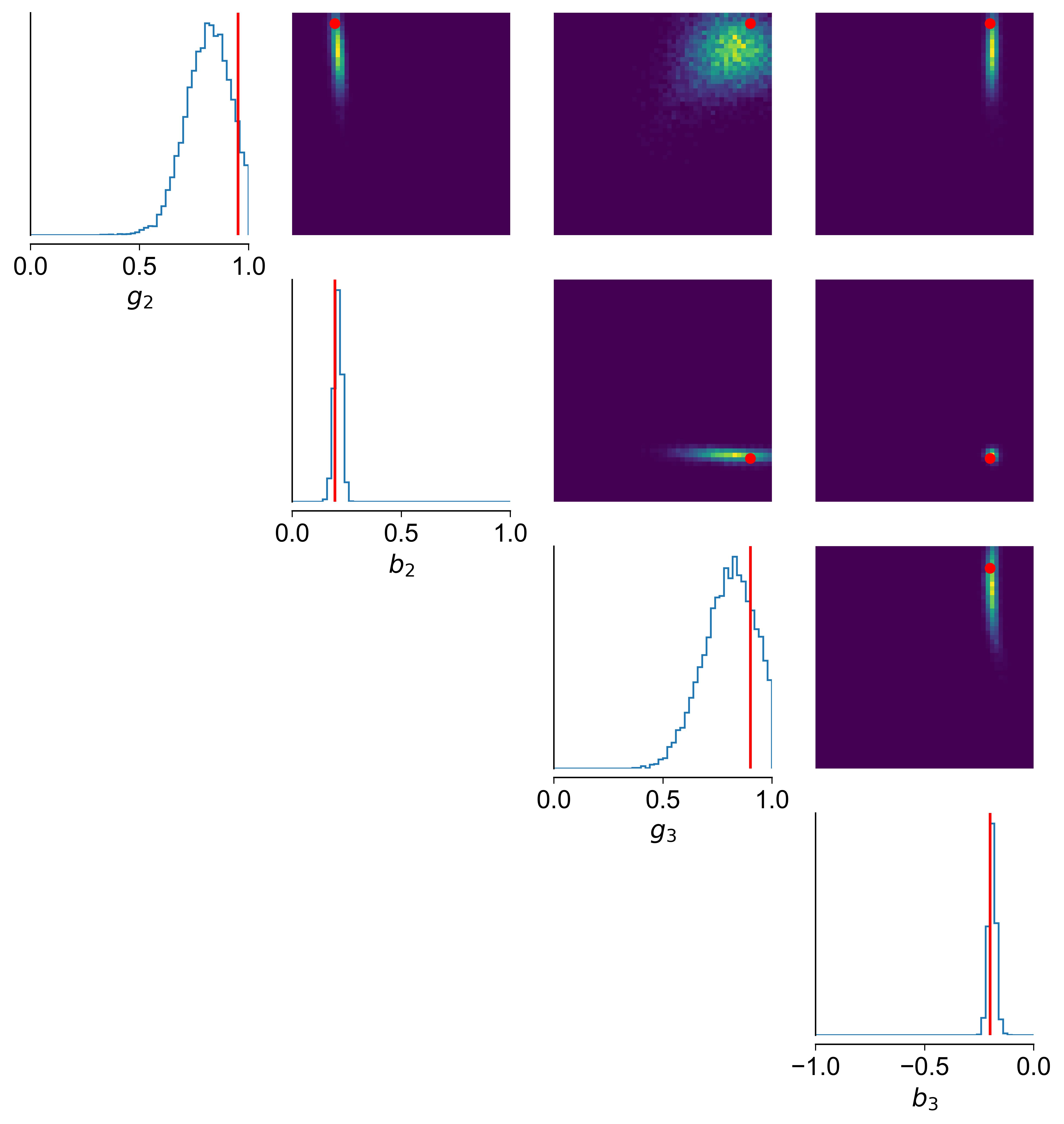}
\caption{GradSubspace-FPA}
\end{subfigure}
\hfill
\begin{subfigure}{0.45\linewidth}
\centering
\includegraphics[width=\linewidth]{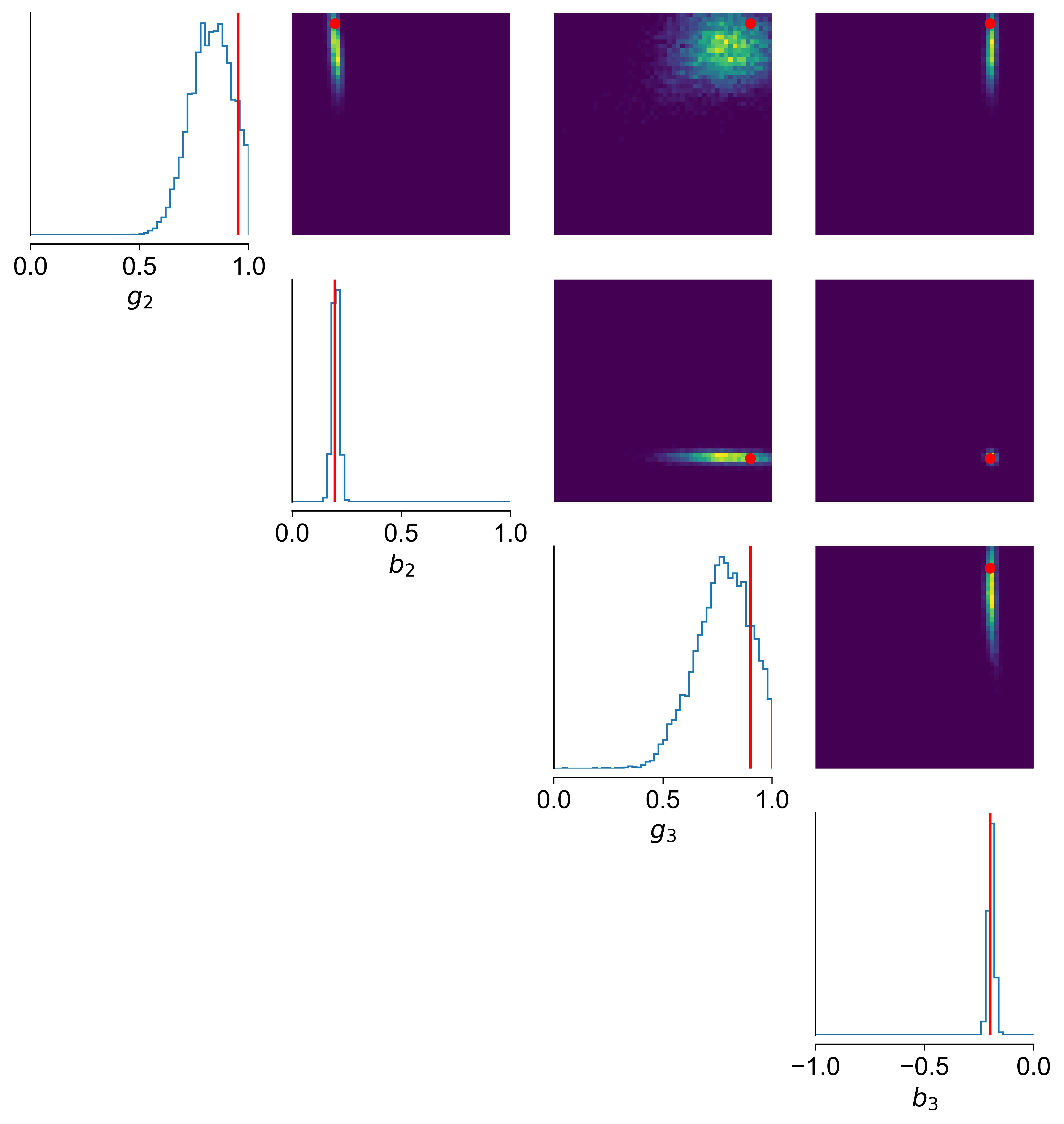}
\caption{GradSubspace-PEA}
\end{subfigure}
\caption{Posterior samples from the four SNPE-TTT strategies for the Brock--Hommes. Diagonal panels show marginal posterior distribution (blue curve), upper-triangular panels show bivariate joint density. Red lines and points denote the mean of the likelihood-based reference posterior. Darker (resp. lighter) colours indicate lower (resp. higher) posterior density in off-diagonal.}
\label{fig:bh_adaptation_posteriors}
\end{figure}

\subsubsection{Results}

Figure~\ref{fig:bh_adaptation_posteriors} shows the posterior inference obtained with SNPE-FPA, SNPE-LoRA, GradSubspace-FPA, and GradSubspace-PEA, respectively. In each subplot, the diagonal panels display marginal posterior distributions, while the upper-triangular panels show bivariate joint distributions. Red lines and points mark the corresponding means of the approximate ground truth posterior sampled from MH.

We observe that the target ground truth posterior in Figure~\ref{fig:pairplotsgt} shows sharply concentrated marginals for $b_2$ and $b_3$, whereas $g_2$ and $g_3$ have broader distributions near the upper prior boundary. This difference in concentration is also visible in the bivariate joint densities. Distributions pairing a $g$ parameter with a $b$ parameter form narrow bands, while the $(b_2,b_3)$ distribution is tightly localized. SNPE-FPA and the two gradient-subspace methods broadly recover these features, particularly the narrow marginals for $b_2$ and $b_3$ near their reference means. Recovery is less precise for $g_2$ and $g_3$, whose marginals are generally broader and shifted towards lower values relative to the ground truth posterior. Correspondingly, their $(g_2,g_3)$ distributions extend further into the lower-value region, although they remain concentrated near the upper part of the prior support. SNPE-LoRA shows a more pronounced mismatch in $b_2$, with its marginal peak shifted towards higher values. This displacement is visible in all three bivariate joint distributions involving $b_2$. Its $g_3$ marginal is also broader than the approximate ground truth distribution. 

\begin{figure}[!htbp]
\centering
\begin{subfigure}{0.96\textwidth}
\centering
\includegraphics[width=\linewidth]{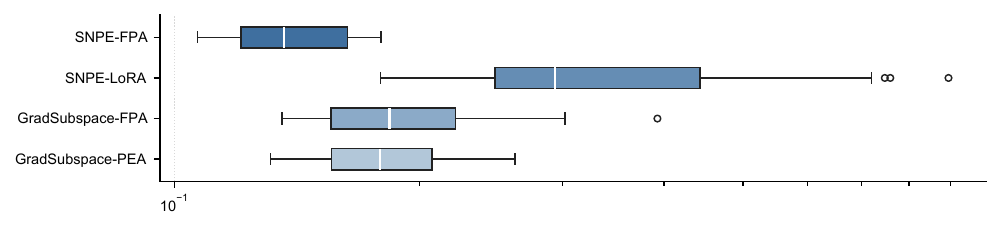}
\caption{Wasserstein distances}
\end{subfigure}
\vspace{0.25em}
\begin{subfigure}{0.96\textwidth}
\centering
\includegraphics[width=\linewidth]{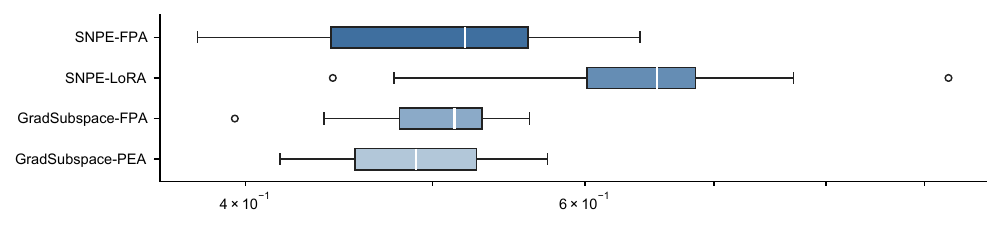}
\caption{Maximum mean discrepancies}
\end{subfigure}
\caption{Boxplots of Brock-Hommes discrepancies over 20 repeated trails for each approach. }
\label{fig:bh_multiseed}
\end{figure}

Additionally, we further explore the robustness of our four approaches by repeating the whole test-time training procedure 20 times at different seeds for the same fixed observation $y$. Figure~\ref{fig:bh_multiseed} and Table~\ref{tab:bh_multiseed} show the posterior inference discrepancies. Under WASS, the SNPE-FPA is located at lower values than those of the gradient-subspace methods, although their interquartile ranges overlap. SNPE-FPA yielded the lowest median WASS of 0.1364, compared with 0.1838 for GradSubspace-FPA and 0.1789 for GradSubspace-PEA. Both gradient-subspace methods had significantly higher WASS values in paired Wilcoxon tests after Holm correction ($p_{\mathrm{Holm}}=0.0006$ and $0.0029$, respectively). SNPE-LoRA shows a higher median of 0.2939, the widest interquartile range, and several observations beyond the upper whisker. The MMD$^2$ boxplots show greater overlap among SNPE-FPA and the two gradient-subspace methods, with both gradient-subspace methods exhibiting narrower interquartile ranges than SNPE-FPA. GradSubspace-PEA had the lowest observed median of 0.4905, followed by GradSubspace-FPA at 0.5133 and SNPE-FPA at 0.5199. However, neither gradient-subspace method differed significantly from SNPE-FPA after Holm correction (both $p_{\mathrm{Holm}}=1.0000$), nor did they differ significantly from each other. SNPE-LoRA again had the highest median, at 0.6537, although its interquartile range was narrower than that of SNPE-FPA. Its comparisons remain descriptive because the offset-seed SNPE-LoRA fits were excluded from the paired tests.

Overall, SNPE-FPA achieved lower WASS than both gradient-subspace methods, whereas the MMD$^2$ comparisons detected no significant differences among these three methods. The tested reduced update spaces therefore showed no consistent posterior-accuracy advantage in this Brock-Hommes setting.

\begin{table}[!htbp]
\centering
\small
\begin{tabular}{lcc}
\toprule
Method & WASS, median [IQR] & MMD$^2$, median [IQR] \\
\midrule
SNPE-FPA & \textbf{0.1364} [0.1207, 0.1631] & 0.5199 [0.4429, 0.5604] \\
SNPE-LoRA & 0.2939 [0.2479, 0.4429] & 0.6537 [0.6016, 0.6846] \\
GradSubspace-FPA & 0.1838 [0.1558, 0.2216] & 0.5133 [0.4808, 0.5307] \\
GradSubspace-PEA & 0.1789 [0.1560, 0.2072] & \textbf{0.4905} [0.4560, 0.5272] \\
\bottomrule
\end{tabular}
\caption{Brock-Hommes posterior discrepancies across 20 repeated trails. Bold values mark the lowest descriptive median in each column.}
\label{tab:bh_multiseed}
\end{table}

\subsection{Multivariate geometric Brownian motion model}

We use multivariate geometric Brownian motion (MVGBM) as an additional tractable example to evaluate our SNPE-TTT. MVGBM is a continuous-time process used in financial time-series modeling, and its discretized transition density is available in closed form. It therefore provides an approximate ground-truth posterior for checking the quality of adapted SNPE-TTT posterior.

We consider a $d$-dimensional positive-valued process $X_t=(X_t^1,\ldots,X_t^d)\in\mathbb{R}^d_{+}$, whose components evolve according to the stochastic differential equation
\begin{equation}
\label{eq:mvgbm_sde}
dX_t^i
=
X_t^i
\left[
b_i\,dt
+
\sum_{j=1}^{d}\sigma_{ij}\, dW_t^j
\right],
\end{equation}
where $b_i$ are drift coefficients, $\sigma_{ij}$ are volatility coefficients, and $\{W_t^j\}_{j=1}^d$ are components of a $d$-dimensional Brownian motion.

In this numerical experiment, we take $d=3$ and infer the posterior of $\theta=(b_1,b_2,b_3)$, given a discretely observed trajectory $y\sim p(x\mid \theta^\ast)$ consisting of $T=100$ equally spaced time points with spacing $\Delta t = 1/(T-1)$. We use independent uniform priors $b_i\sim \mathcal{U}(-1,1)$ for $i=1,2,3$.

Form Equation~\ref{eq:mvgbm_sde}, we can obtain the likelihhod od MVGBM by a log transformation. Let $Z_t=\log X_t$. Then
\begin{equation}
\label{eq:mvgbm_transition}
X_{t+\Delta t}=\exp Z_{t+\Delta t},
\qquad
Z_{t+\Delta t}\sim
\mathcal{N}\!\Bigl(
\log X_t + (\theta-\gamma)\Delta t;
\sigma\sigma^\top \Delta t
\Bigr),
\end{equation}
where $\exp(\cdot)$ denotes elementwise exponentiation and
\begin{equation}
\label{eq:mvgbm_gamma_sigma}
\gamma
=
\frac{1}{2}
\left[
\sum_{j=1}^{d}\sigma_{1j}^{2},
\sum_{j=1}^{d}\sigma_{2j}^{2},
\sum_{j=1}^{d}\sigma_{3j}^{2}
\right]^{\!\top},
\qquad
\sigma=
\begin{pmatrix}
0.5 & 0.1 & 0.0\\
0.0 & 0.1 & 0.3\\
0.0 & 0.0 & 0.2
\end{pmatrix}.
\end{equation}

Because the transition density in Eq.~\eqref{eq:mvgbm_transition} is available, we evaluate the likelihood numerically and obtain reference samples using MH. More implementation convention is specified in Appendix~\ref{app:experiments}, and posterior sampling in Appendix~\ref{app:references}.

\subsubsection{Model Regime Shift}

\begin{figure}[!htbp]
\centering
\begin{subfigure}{0.45\linewidth}
\centering
\includegraphics[width=\linewidth]{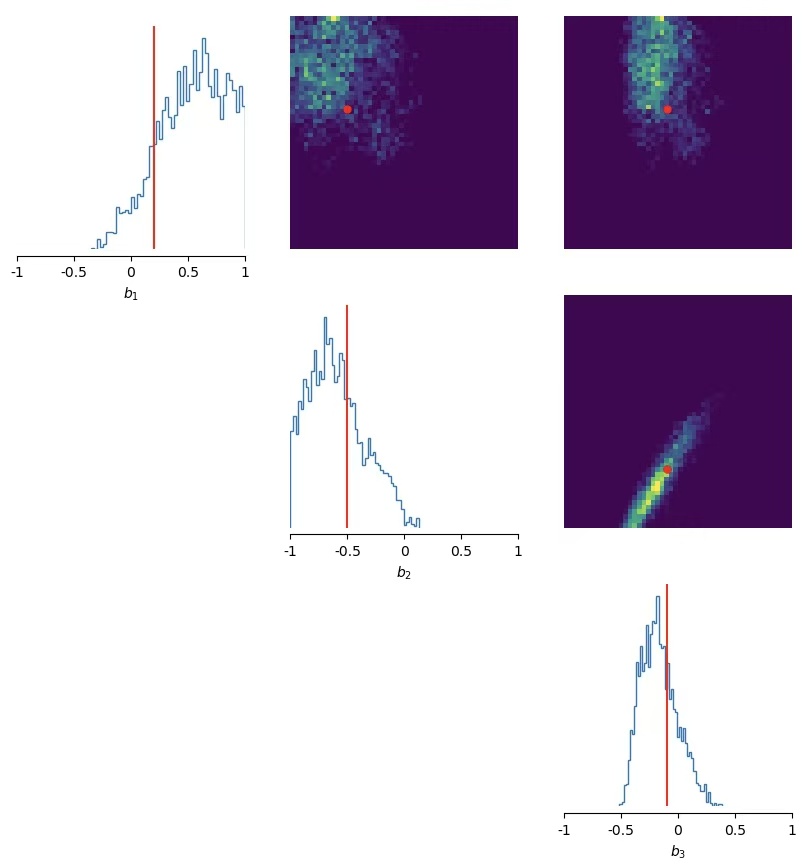}
\caption{Source model}
\end{subfigure}
\hfill
\begin{subfigure}{0.45\linewidth}
\centering
\includegraphics[width=\linewidth]{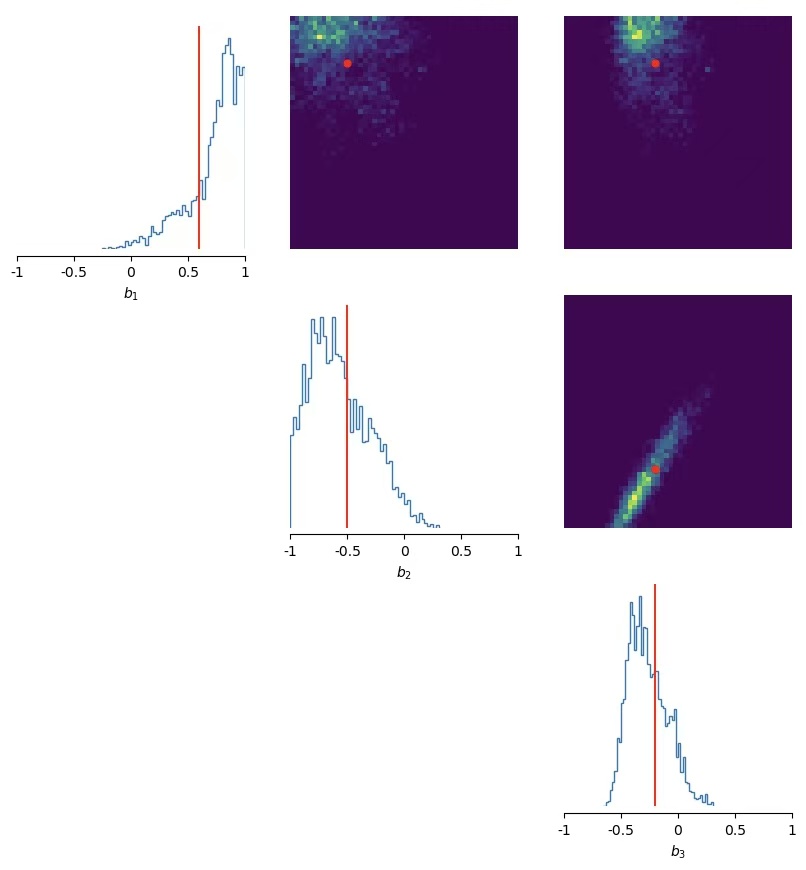}
\caption{Target model}
\end{subfigure}
\caption{Approximate ground-truth posteriors for the MVGBM source and target models.}
\label{fig:gtmvgbm}
\end{figure}

We first train a base SNPE estimator under MVGBM using observations generated at $\theta^\ast=(0.2,-0.5,-0.1)$, and then adapt it using simulations for a new target observation generated at $\theta^\ast_{\mathrm{new}}=(0.6,-0.5,-0.2)$ and other settings unchanged. Consequently, this experiment studies a parameter-induced change in the observed trajectory and posterior location rather than a simulator or mechanism shift. Figure~\ref{fig:gtmvgbm} shows the likelihood-based reference posteriors for the source and target observations.

\subsubsection{Results}

\begin{figure}[!htbp]
\centering
\begin{subfigure}{0.45\linewidth}
\centering
\includegraphics[width=\linewidth]{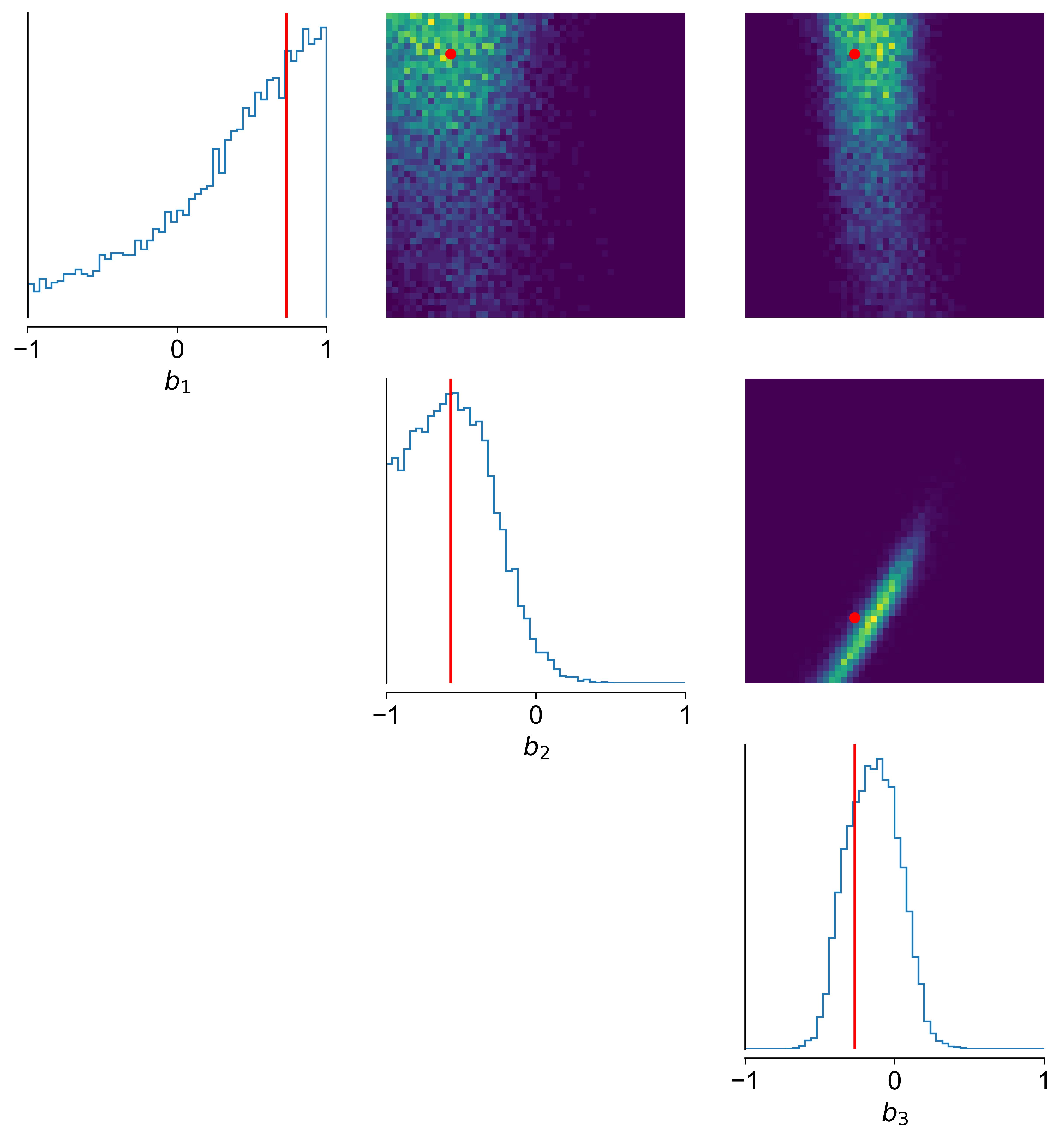}
\caption{SNPE-FPA}
\end{subfigure}
\hfill
\begin{subfigure}{0.45\linewidth}
\centering
\includegraphics[width=\linewidth]{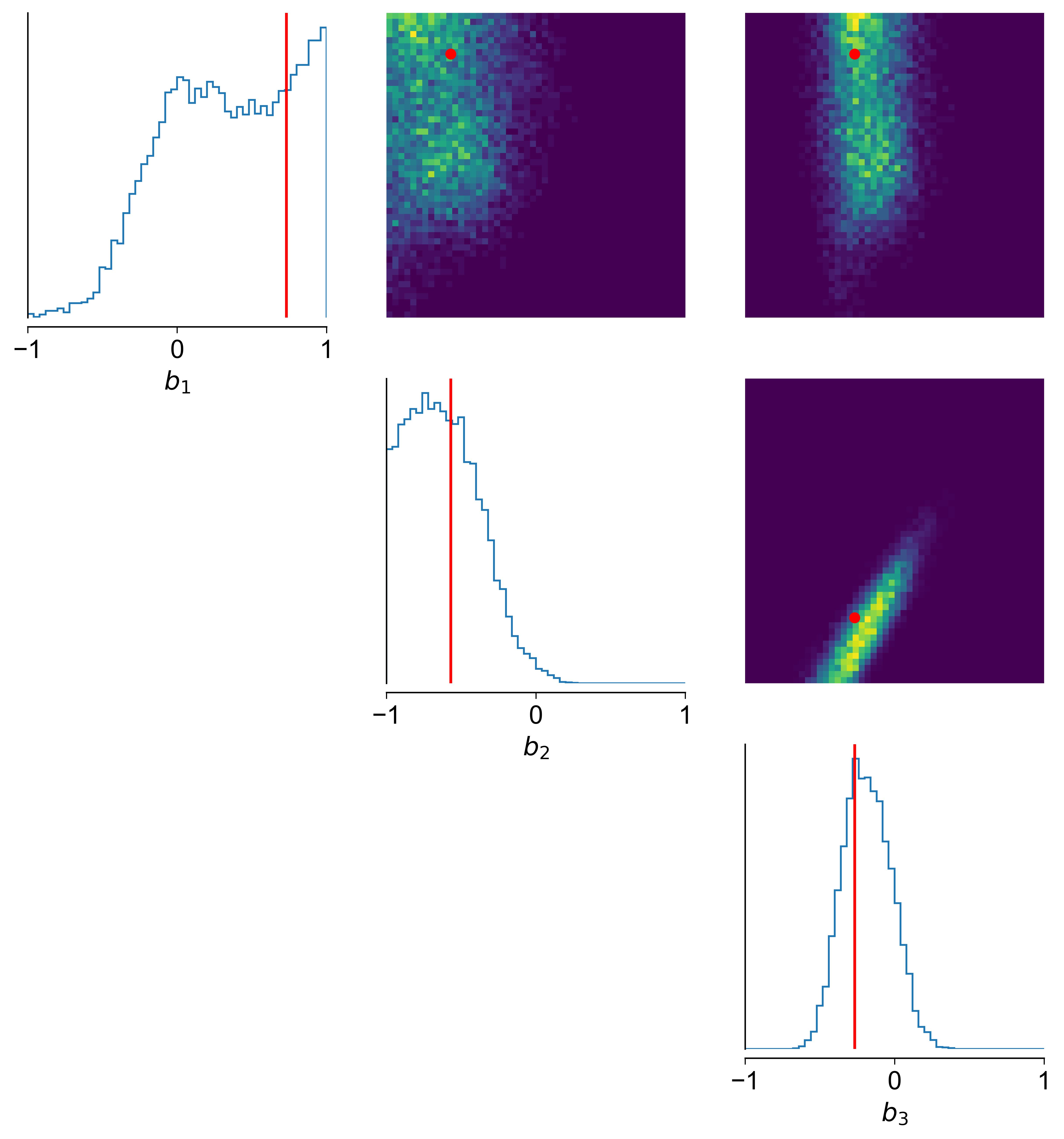}
\caption{SNPE-LoRA}
\end{subfigure}
\begin{subfigure}{0.45\linewidth}
\centering
\includegraphics[width=\linewidth]{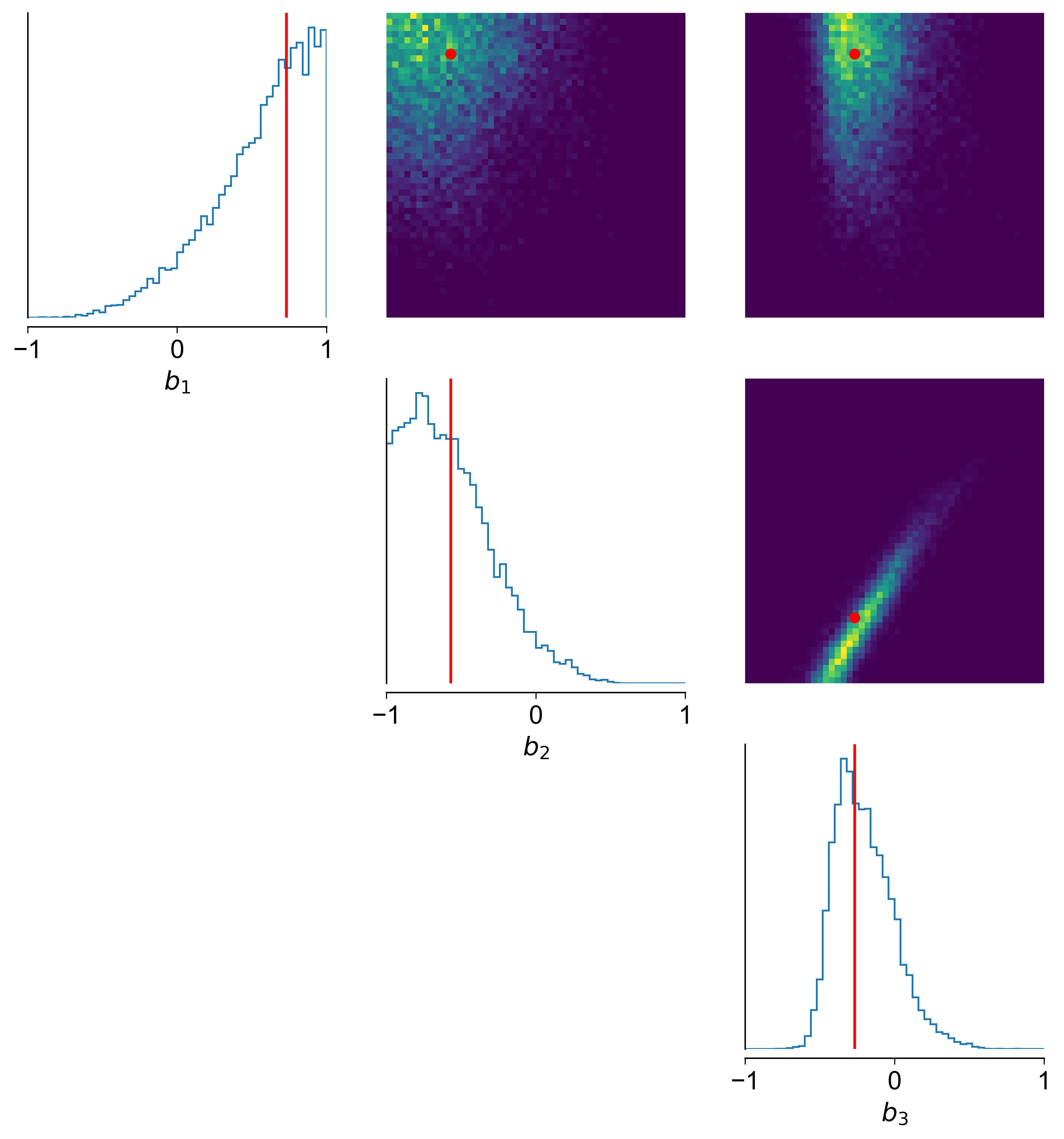}
\caption{GradSubspace-FPA}
\end{subfigure}
\hfill
\begin{subfigure}{0.45\linewidth}
\centering
\includegraphics[width=\linewidth]{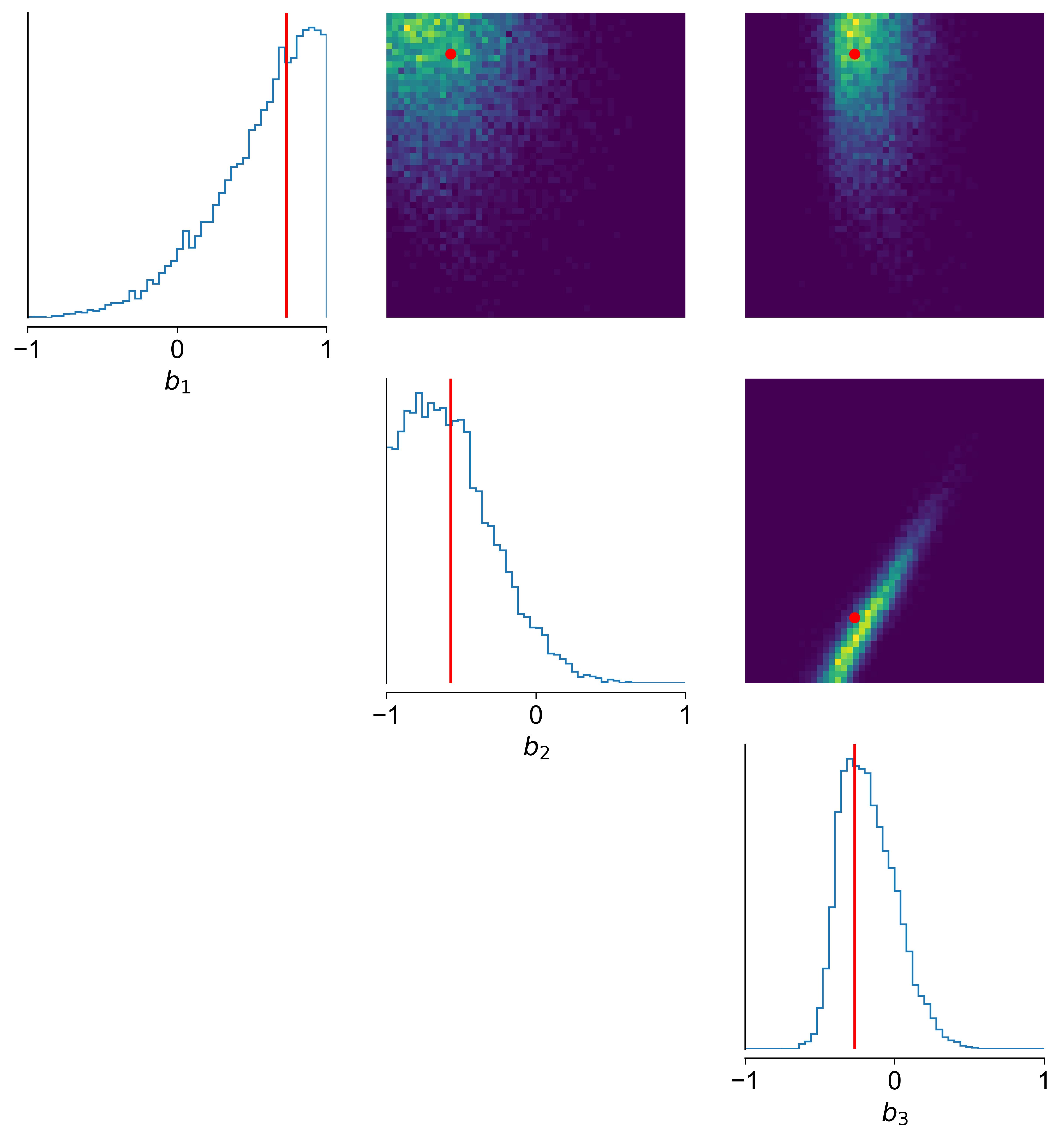}
\caption{GradSubspace-PEA}
\end{subfigure}
\caption{Posterior samples from the four SNPE-TTT strategies for the MVGBM. Diagonal panels show marginal posterior distribution (blue curve), upper-triangular panels show bivariate joint density. Red lines and points denote the mean of the likelihood-based reference posterior. Darker (resp. lighter) colours indicate lower (resp. higher) posterior density in off-diagonal.}
\label{fig:mv_adaptation_posteriors}
\end{figure}

The MVGBM experiment follows the same simulation budget, repeated-fit protocol, discrepancy metric, and statistical analysis as the Brock-Hommes experiment. Figure~\ref{fig:mv_adaptation_posteriors} compares posterior samples produced by the four adapted SNPE estimators with the approximate ground-truth posterior. All four methods reproduce the strong positive dependence between $b_2$ and $b_3$, although differences remain in marginal location, concentration, and dispersion. These differences are particularly visible in the shape of the $b_1$ marginal. 

\begin{figure}[!htbp]
\centering
\begin{subfigure}{0.96\textwidth}
\centering
\includegraphics[width=\linewidth]{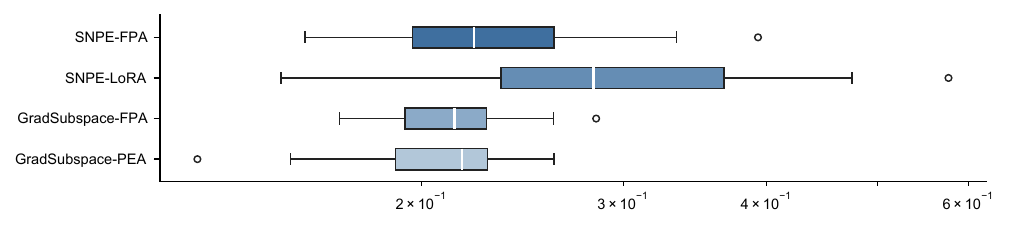}
\caption{Wasserstein distances}
\end{subfigure}
\vspace{0.25em}
\begin{subfigure}{0.96\textwidth}
\centering
\includegraphics[width=\linewidth]{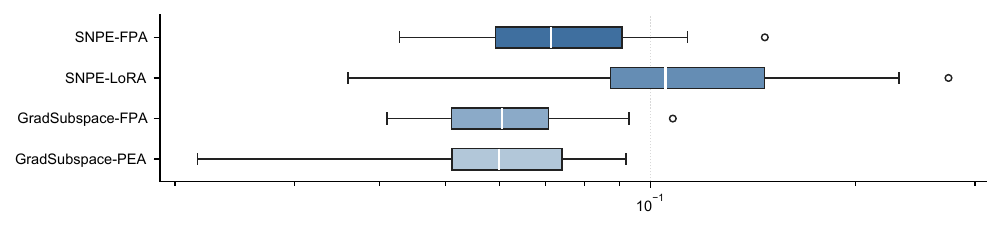}
\caption{Maximum mean discrepancies}
\end{subfigure}
\caption{Boxplots of MVGBM discrepancies over 20 repeated trails for each approach.}
\label{fig:mv_multiseed}
\end{figure}

In Figure~\ref{fig:mv_multiseed} and Table~\ref{tab:mv_multiseed}, we show the posterior discrepancies between our adapted SNPE-TTT posterior inference and approximate ground truth over 20 repeated fits for each method. GradSubspace-FPA and GradSubspace-PEA had median WASS values of 0.2137 and 0.2169, respectively, compared with 0.2221 for SNPE-FPA. Although both descriptive medians were lower than that of SNPE-FPA, neither difference remained significant after Holm correction ($p_{\mathrm{Holm}}=0.4049$ and $0.3982$, respectively). For MMD$^2$, the median values were 0.0604 for GradSubspace-FPA, 0.0598 for GradSubspace-PEA, and 0.0715 for SNPE-FPA. The difference between GradSubspace-PEA and SNPE-FPA remained significant after Holm correction ($p_{\mathrm{Holm}}=0.0283$), whereas the GradSubspace-FPA comparison did not ($p_{\mathrm{Holm}}=0.1165$). GradSubspace-FPA and GradSubspace-PEA did not differ significantly under either discrepancy measure. SNPE-LoRA again had higher descriptive medians and wider interquartile ranges than the other three methods.

\begin{table}[!htbp]
\centering
\small
\begin{tabular}{lcc}
\toprule
Method & WASS, median [IQR] & MMD$^2$, median [IQR] \\
\midrule
SNPE-FPA & 0.2221 [0.1964, 0.2611] & 0.0715 [0.0592, 0.0908] \\
SNPE-LoRA & 0.2825 [0.2345, 0.3672] & 0.1053 [0.0874, 0.1470] \\
GradSubspace-FPA & \textbf{0.2137} [0.1934, 0.2278] & 0.0604 [0.0510, 0.0708] \\
GradSubspace-PEA & 0.2169 [0.1898, 0.2284] & \textbf{0.0598} [0.0511, 0.0742] \\
\bottomrule
\end{tabular}
\caption{MVGBM posterior discrepancies across 20 repeated trails. Bold values mark the lowest descriptive median in each column.}
\label{tab:mv_multiseed}
\end{table}

\FloatBarrier
\section{Numerical Experiments for intractable example}
\label{sec:fw_experiments}

In pervious section, we showed that SNPE-TTT framework have a demonstrated performance in posterior inference of likelihood-tractable ABMs, with both more accurate and simulation-efficient. We next evaluate the four SNPE-TTT adaptation approaches in an ABM for which a likelihood is unavailable. We use the Franke-Westerhoff financial-market model and assess simulation-based calibration across independently generated observations. Further experimental details and calibration results are provided in Appendix~\ref{app:fw_sbc}.

\subsection{Simulation-Based Calibration}
\label{sec:fw_metrics}

In Section~\ref{sec:experiments}, we evaluated the adapted posteriors using WASS and MMD$^2$, both of which require samples from an approximate ground-truth posterior. In practice, however, neither the likelihood nor a reliable reference posterior is always available, as in the likelihood-intractable Franke-Westerhoff model considered in this section. We therefore employ simulation-based calibration (SBC), a widely used diagnostic for assessing the calibration of approximate Bayesian inference procedures without requiring reference posterior samples \citep{talts2018sbc}.

The core idea for SBC follows from the fact that, under exact Bayesian inference, the posterior averaged over the prior predictive distribution recovers the prior:
\begin{equation}
\int_{\mathcal{Y}}\int_{\Theta}
p(\theta \mid \tilde{y})\,
p(\tilde{y} \mid \tilde{\theta})\,
p(\tilde{\theta})
\,\mathrm{d}\tilde{\theta}\,\mathrm{d}\tilde{y}
=
p(\theta).
\label{eq:sbc_identity}
\end{equation}
This identity motivates the use of prior predictive simulations to assess posterior calibration. Rather than directly comparing the data-averaged approximate posterior with the prior, SBC examines the ranks of generating parameters among samples from their corresponding approximate posteriors \citep{talts2018sbc,modrak2025sbc}.

Following \citet{dyer2024black}, we conduct SBC across $M$ independent experiments. For each experiment $m=1,\ldots,M$, we proceed as follows:
\begin{enumerate}
    \item Draw a generating parameter from the prior, $\tilde{\theta}_m \sim p(\theta)$.
    \item Simulate a synthetic observation from the model, $\tilde{y}_m \sim p(y \mid \tilde{\theta}_m)$.
    \item Draw $J$ independent samples from the approximate posterior conditioned on $\tilde{y}_m$, $\theta_m^{(j)} \sim q_{\phi}(\theta \mid \tilde{y}_m)$, for $j=1,\ldots,J$.
    \item For each parameter component $k$, compute the rank statistic
    \begin{equation}
    r_{mk}
    =
    \sum_{j=1}^{J}
    \mathbf{1}\!\left\{
    \theta_{mk}^{(j)} < \tilde{\theta}_{mk}
    \right\},
    \qquad
    r_{mk}\in\{0,\ldots,J\}.
    \label{eq:sbc_rank}
    \end{equation}
\end{enumerate}
Under exact posterior inference, the generating parameter and posterior samples are exchangeable conditional on the simulated observation. Consequently, the rank statistics follow a discrete uniform distribution on $\{0,\ldots,J\}$ for each parameter component \citep{talts2018sbc}. Systematic deviations from uniformity therefore provide evidence of posterior miscalibration.

In our implementation, we use $J=999$ posterior samples, partitioning the resulting $J+1=1000$ possible ranks into $B=10$ equally sized bins. Each rank histogram displays the expected bin count $M/B$ as a solid black line, together with a central $95\%$ binomial interval shown as a grey band. Under exact posterior inference, the count in each bin follows a $\operatorname{Binomial}(M,1/B)$ distribution. A flat rank histogram is consistent with marginal posterior calibration, whereas U-shaped, skewed, or otherwise structured histograms may indicate underdispersion, overdispersion, bias, or other systematic distortions in the approximate posterior. We interpret SBC as a complementary diagnostic rather than a definitive measure of posterior accuracy. For the Franke-Westerhoff model, SBC provides evidence about marginal calibration but does not establish the overall accuracy of the posterior inference. Moreover, for our trained conditional density estimators, posterior samples can be generated directly without additional simulator calls during posterior sampling, making SBC a practical diagnostic for likelihood-intractable models.

\subsection{Franke--Westerhoff Model}
\label{sec:fw_model}

We consider the wealth-and-predisposition version of the Franke-Westerhoff model \citep{franke2012structural}. The market contains fundamentalist traders, who respond to deviations from a fundamental price, and chartist traders, who follow recent price changes. Let $p_t$ denote the log price and $p^\star$ its fundamental value. The demands of the two groups are
\begin{align}
d_t^f &= \phi(p^\star-p_t)+\sigma_f\epsilon_t^f,
\label{eq:fw_fundamentalist}\\
d_t^c &= \chi(p_t-p_{t-1})+\sigma_c\epsilon_t^c,
\label{eq:fw_chartist}
\end{align}
where $\epsilon_t^f$ and $\epsilon_t^c$ are independent standard normal shocks. Their realised gains and smoothed wealth measures satisfy
\begin{align}
g_t^j &= (e^{p_t}-e^{p_{t-1}})d_{t-2}^j,
\qquad j\in\{f,c\},
\label{eq:fw_gains}\\
w_t^j &= \eta w_{t-1}^j+(1-\eta)g_t^j.
\label{eq:fw_wealth}
\end{align}
The relative attractiveness of fundamentalism is $a_t=\alpha_w(w_t^f-w_t^c)+\alpha_0$. The population fractions and price update are
\begin{align}
n_t^f &= \frac{1}{1+\exp(-\beta a_{t-1})},
\qquad n_t^c=1-n_t^f,
\label{eq:fw_shares}\\
p_{t+1} &= p_t+\mu\left(n_t^f d_t^f+n_t^c d_t^c\right).
\label{eq:fw_price}
\end{align}
Thus, $\alpha_w$ determines the influence of relative wealth on strategy choice, $\eta$ controls wealth persistence, and $\sigma_c$ determines the chartist demand-noise scale. The switching intensity $\beta$ governs the sensitivity of the population fractions to relative attractiveness.

The observed output is the log-return series $r_t=p_t-p_{t-1}$. We infer $\theta=(\alpha_w,\eta,\sigma_c)$ under independent priors $\alpha_w\sim\mathcal U(0,15000)$, $\eta\sim\mathcal U(0,1)$ and $\sigma_c\sim\mathcal U(0,5)$. Fixed values are $\mu=0.01$, $\phi=1$, $\chi=0.9$, $\alpha_0=2.1$, $\sigma_f=0.752$ and $p^\star=0$. Unlike the tractable examples, the FW experiment does not provide an exact likelihood evaluation, so we cannot construct an MH approximate ground-truth posterior for this task.

\subsubsection{Model Regime Shift}
\label{sec:fw_regime}

For our experiment in Franke-Westerhoff Model, we set the source simulator using $\beta_0=0.8$ and the target simulator using $\beta_1=1$, with the same priors and remaining fixed coefficients. Equation~\eqref{eq:fw_shares} gives $\partial n_t^f/\partial\beta=a_{t-1}n_t^f(1-n_t^f)$ when attractiveness is held fixed. Increasing $\beta$ therefore strengthens selection towards whichever strategy has positive relative attractiveness, and it does not uniformly increase the fundamentalist share. As attractiveness depends on previous wealth and prices, this change can affect both the return dynamics and the target inference problem.

The source estimator was trained for ten SNPE rounds of 1,000 simulations around one independently generated source observation. Every target fit reused the same pre-trained SNPE. Target adaptation used four rounds with budgets $[500,500,500,1000]$, totalling 2,500 target simulations which is only one-quarter of the original pre-trained SNPE. The first target proposal was the prior, and subsequent proposals were the preceding target posterior, with SNPE-C proposal correction. 

\begin{figure}[!htbp]
\centering
\begin{subfigure}[t]{0.45\linewidth}
\centering
\includegraphics[width=\linewidth]{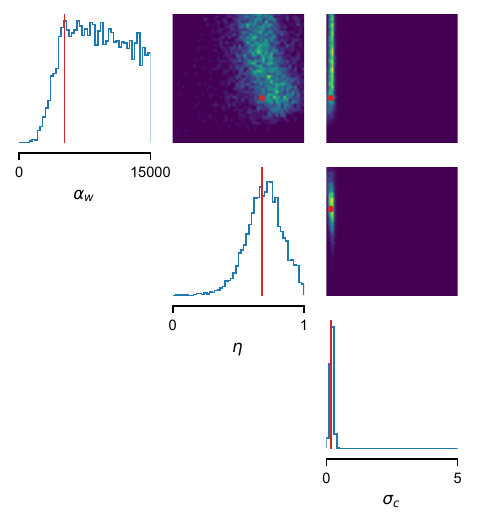}
\caption{Source SNPE inference ($\beta=0.8$)}
\end{subfigure}
\hfill
\begin{subfigure}[t]{0.45\linewidth}
\centering
\includegraphics[width=\linewidth]{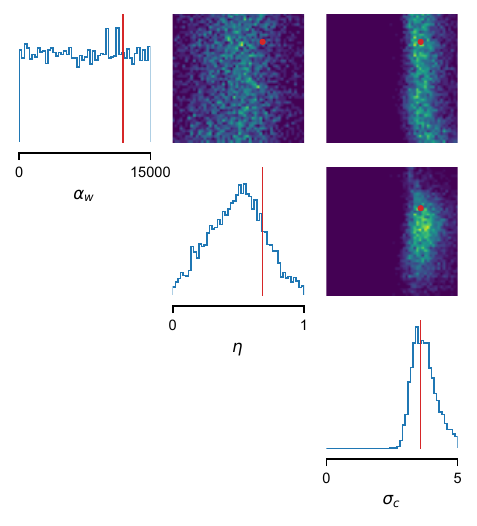}
\caption{Target SNPE inference( $\beta=1$)}
\end{subfigure}
\caption{Posterior samples from SNPE-TTT for the source (left) and target (right) Franke-Westerhoff Model. Diagonal panels show marginal posterior distribution (blue curve), upper-triangular panels show bivariate joint density. Red lines and points indicate the generating parameter values. Darker (resp. lighter) colours indicate lower (resp. higher) posterior density in off-diagonal.}
\label{fig:fw_regime_posteriors}
\end{figure}

Figure~\ref{fig:fw_regime_posteriors} shows the SNPE posterior and the target posterior of Franke-Westerhoff, which are learned by SNPE. Consequently, differences between these distribution reflect changes in observation and generating parameters as well as the change in $\beta$ and the different training budgets. 

\subsubsection{Results}
\label{sec:fw_results}

We evaluate SNPE-TTT using 100 independent target observations drawn from the prior predictive distribution. All methods are evaluated using the same observations to ensure a consistent comparison. Further implementation details and additional results are provided in Appendix~\ref{app:fw_sbc}.

In Figure~\ref{fig:fw_adaptation_posteriors}, we compares the posterior distributions obtained using the four SNPE-TTT adaptation strategies for the same target observation shown in Figure~\ref{fig:fw_regime_posteriors}(b). Across the four methods, the $\alpha_w$ marginals remain relatively broad, whereas the $\sigma_c$ marginals are more concentrated. Differences are also observed in the shape and location of the $\eta$ marginal, indicating that the adaptation strategies produce different posterior approximations for the same observation. Although the generating-parameter markers provide a useful reference, their agreement with regions of high posterior density does not necessarily imply accurate posterior inference. This comparison therefore provides a qualitative illustration of the inferred posteriors, while their calibration across repeated observations is assessed using SBC.

\begin{figure}[!htbp]
\centering
\begin{subfigure}[t]{0.9\linewidth}
\centering
\includegraphics[width=\linewidth]{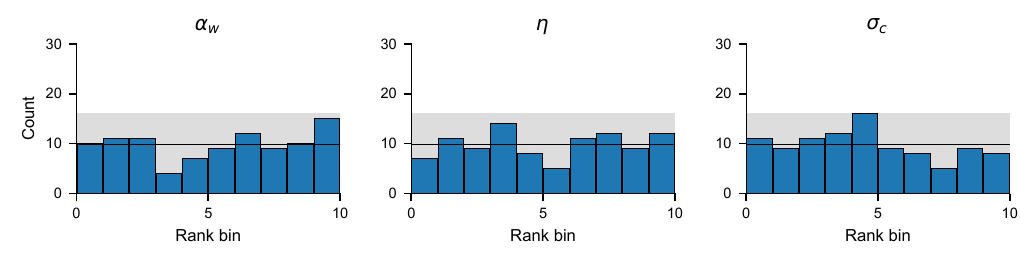}
\caption{SNPE-FPA}
\end{subfigure}
\par\smallskip
\begin{subfigure}[t]{0.9\linewidth}
\centering
\includegraphics[width=\linewidth]{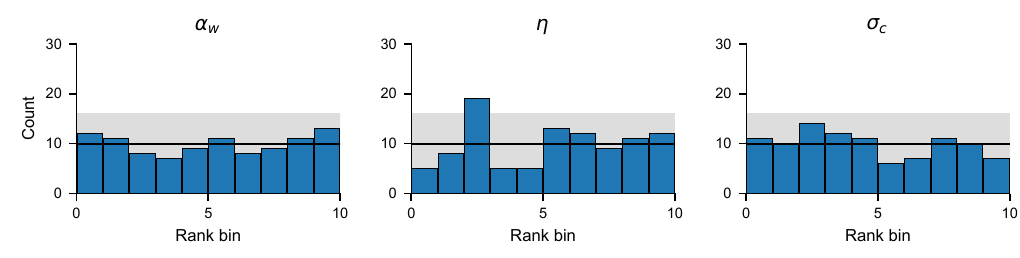}
\caption{SNPE-LoRA}
\end{subfigure}
\par\smallskip
\begin{subfigure}[t]{0.9\linewidth}
\centering
\includegraphics[width=\linewidth]{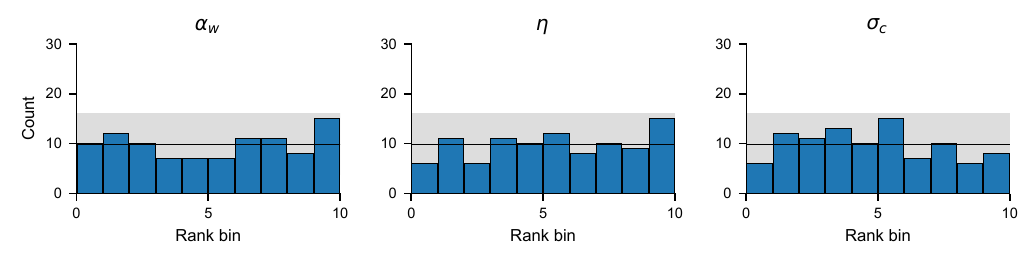}
\caption{GradSubspace-FPA}
\end{subfigure}
\par\smallskip
\begin{subfigure}[t]{0.9\linewidth}
\centering
\includegraphics[width=\linewidth]{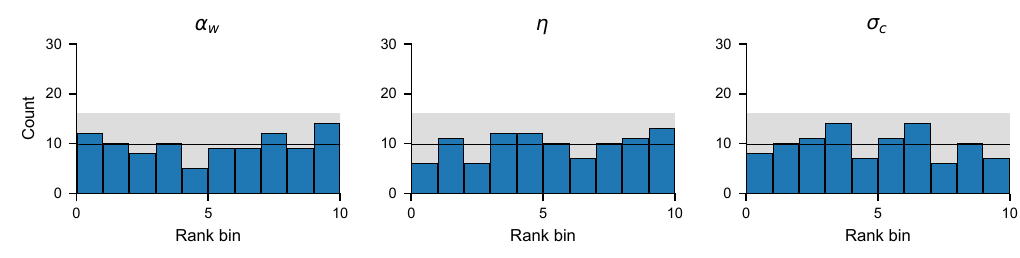}
\caption{GradSubspace-PEA}
\end{subfigure}
\caption{Rank histograms generated according to the simulation-based calibration for SNPE-FPA, SNPE-LoRA, GradSubspace-FPA and GradSubspace-PEA in posterior inference of FW model, respectively. The shaded regions represent the symmetric 95\% point function of a Binomial random variable with success probability 1/P=1/10, that is, the bottom (resp. top) of the shaded regions is the inverse of the cumulative distribution function for a Binomial random variable at quantile 0.025 (resp. 0.975). This generally stands for the expected deviations from complete uniformity when the histograms are distributed uniformly.}
\label{fig:fw_sbc_adaptation}
\end{figure}

In Figure~\ref{fig:fw_sbc_adaptation}, we presents the marginal SBC rank histograms for the four adaptation strategies. Overall, most bin counts remain close to the uniform reference, although an isolated peak is observed in the $\eta$ histogram for SNPE-LoRA. For comparison, Figure~\ref{fig:fw_sbc_baselines} presents the SBC histograms for the baseline methods, with further results provided in Appendix~\ref{app:fw_sbc}. A more pronounced deviation is observed for source-only SNPE in $\sigma_c$, where the concentration of low ranks indicates that the generating values frequently fall below the posterior samples, suggesting a tendency to overestimate $\sigma_c$ across the evaluated observations. In contrast, the adaptation methods exhibit rank distributions that are broadly consistent with uniformity. These findings suggest that SNPE-TTT are able to improve marginal posterior calibration relative to direct source-only SNPE inference with a smaller simulation budget.

\begin{figure}[!htbp]
\centering
\begin{subfigure}[t]{0.45\linewidth}
\centering
\includegraphics[width=\linewidth]{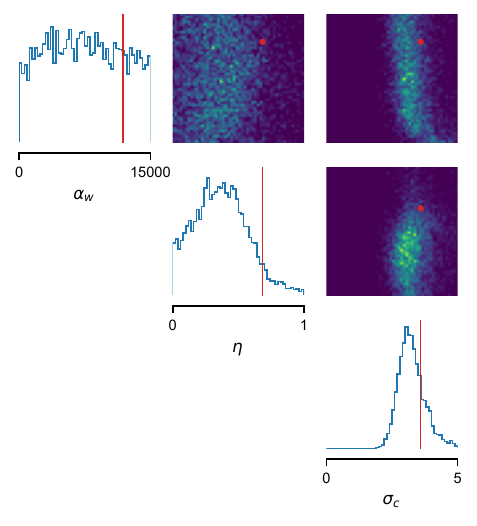}
\caption{SNPE-FPA}
\end{subfigure}
\hfill
\begin{subfigure}[t]{0.45\linewidth}
\centering
\includegraphics[width=\linewidth]{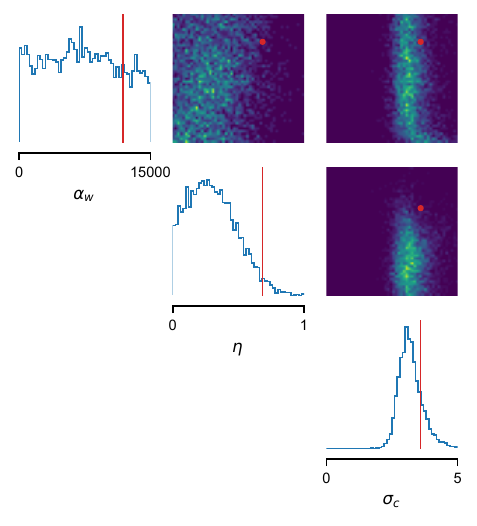}
\caption{SNPE-LoRA}
\end{subfigure}
\par\medskip
\begin{subfigure}[t]{0.45\linewidth}
\centering
\includegraphics[width=\linewidth]{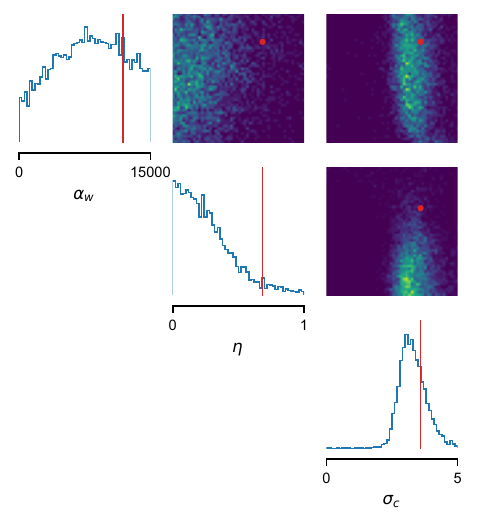}
\caption{GradSubspace-FPA}
\end{subfigure}
\hfill
\begin{subfigure}[t]{0.45\linewidth}
\centering
\includegraphics[width=\linewidth]{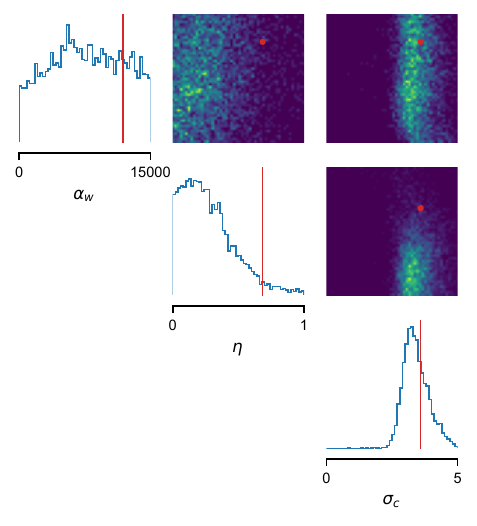}
\caption{GradSubspace-PEA}
\end{subfigure}
\caption{Posterior samples from the four SNPE-TTT strategies for the Franke-Westerhoff Model. Diagonal panels show marginal posterior distribution (blue curve), upper-triangular panels show bivariate joint density. Red lines and points indicate the generating parameter values. Darker (resp. lighter) colours indicate lower (resp. higher) posterior density in off-diagonal.}
\label{fig:fw_adaptation_posteriors}
\end{figure}

\FloatBarrier
\section{Conclusion}
\label{sec:discussion}

In this paper, we investigated SNPE-TTT framework to adapt pre-trained neural posterior estimators to related inference tasks in ABMs. Within this framework, we compared four adaptation strategies: full-parameter SNPE-FPA, SNPE-LoRA, GradSubspace-FPA, and GradSubspace-PEA, which share a sequential simulation and training procedure but differ in the space of allowable parameter updates. We argue that SNPE-TTT is a promising approach to simulation-based inference for parameter posterior estimation in ABMs, particularly those exhibiting complex and stochastic non-equilibrium dynamics, where the high computational cost of simulations makes efficient posterior estimation especially important.

We further benchmarked four SNPE-TTT strategies on both likelihood-tractable and likelihood-intractable ABMs under changes in inference tasks. In the likelihood-tractable benchmarks, SNPE-TTT generally recovered the approximate reference posteriors using only one-quarter of simulation budget required by SNPE trained from scratch. However, the relative performance of different update strategies depended on the model and evaluation metric. Full-parameter adaptation achieved lower WASS than the gradient-subspace methods in Brock-Hommes, while GradSubspace-PEA achieved lower MMD$^2$ than full-parameter adaptation in MVGBM. Although restricted adaptation reduced trainable parameters or optimizer memory, these savings did not consistently translate into improved posterior accuracy. In the likelihood-intractable Franke-Westerhoff model, simulation-based calibration provided additional evidence of posterior reliability. Full-parameter adaptation also achieved shorter target-stage training times than SNPE trained from scratch on successfully evaluated observations, whereas restricted adaptation methods were generally slower under the chosen training settings.

We also acknowledge that there are limitations of SNPE-TTT and thus suggest directions for future research. First, restricting parameter updates may exclude directions important for posterior recovery. Although the gradient-subspace methods retained most of the sampled gradient energy, this did not consistently improve posterior accuracy. Future work might investigate adaptive rank selection and dynamic basis updates to better balance posterior accuracy and computational efficiency. Second, reducing the number of trainable parameters does not necessarily reduce overall computational cost. SNPE-TTT still requires target simulations, and additional optimization overhead may offset the benefits of restricted adaptation. Future studies should therefore evaluate computational efficiency more comprehensively, accounting for simulation costs, memory requirements, and total training time, including source pre-training.

Overall, our results suggest that SNPE-TTT offers a promising approach to simulation-based parameter inference for ABMs, particularly when simulations are computationally expensive and previously trained posterior estimators are available. Nevertheless, the choice of adaptation strategy should balance posterior accuracy, simulation requirements, and computational costs.

\bibliographystyle{apalike}
\bibliography{sample.bib}
\clearpage
\appendix
\section{Further details on SNPE}
\label{app:snpe}
\label{app:implementation}

\subsection{Network architecture and sequential training}

For SNPE posterior inference as mentioned in Secion~\ref{sec:experiments} and \ref{sec:fw_experiments}, we use a two-layer GRU with hidden size 32 and a 16-dimensional context. The conditional Masked Autoregressive Flow (MAF) has five transforms and two autoregressive blocks per transform and 50 hidden features per block. These MAF and optimization settings follow the implementation choices in \citet{dyer2024black}. 

Source SNPE inference use 1,000 simulations in each ten rounds for BH at $\beta=120$, MVGBM at $b^\ast=(0.2,-0.5,-0.1)$, and FW at $\beta=0.8$. SNPE-TTT adaptation for target fits accumulate simulations over four rounds with budgets $[500,500,500,1000]$. The first proposal is the prior; later proposals are the preceding posterior conditioned on the target observation. Training uses SNPE-C proposal correction, learning rate $5\times10^{-4}$, batches of 50, a 10\% validation split and gradient clipping at norm 5. Weights carry over between rounds, while Adam moments are reset. Early stopping uses patience 20 and FW additionally imposes a 300-epoch cap. 

\subsection{SNPE posterior visualisations for BH and MVGBM}
\label{app:snpe_posteriors}

\begin{figure}[!htbp]
\centering
\begin{subfigure}[t]{0.48\textwidth}
\centering
\includegraphics[width=\linewidth]{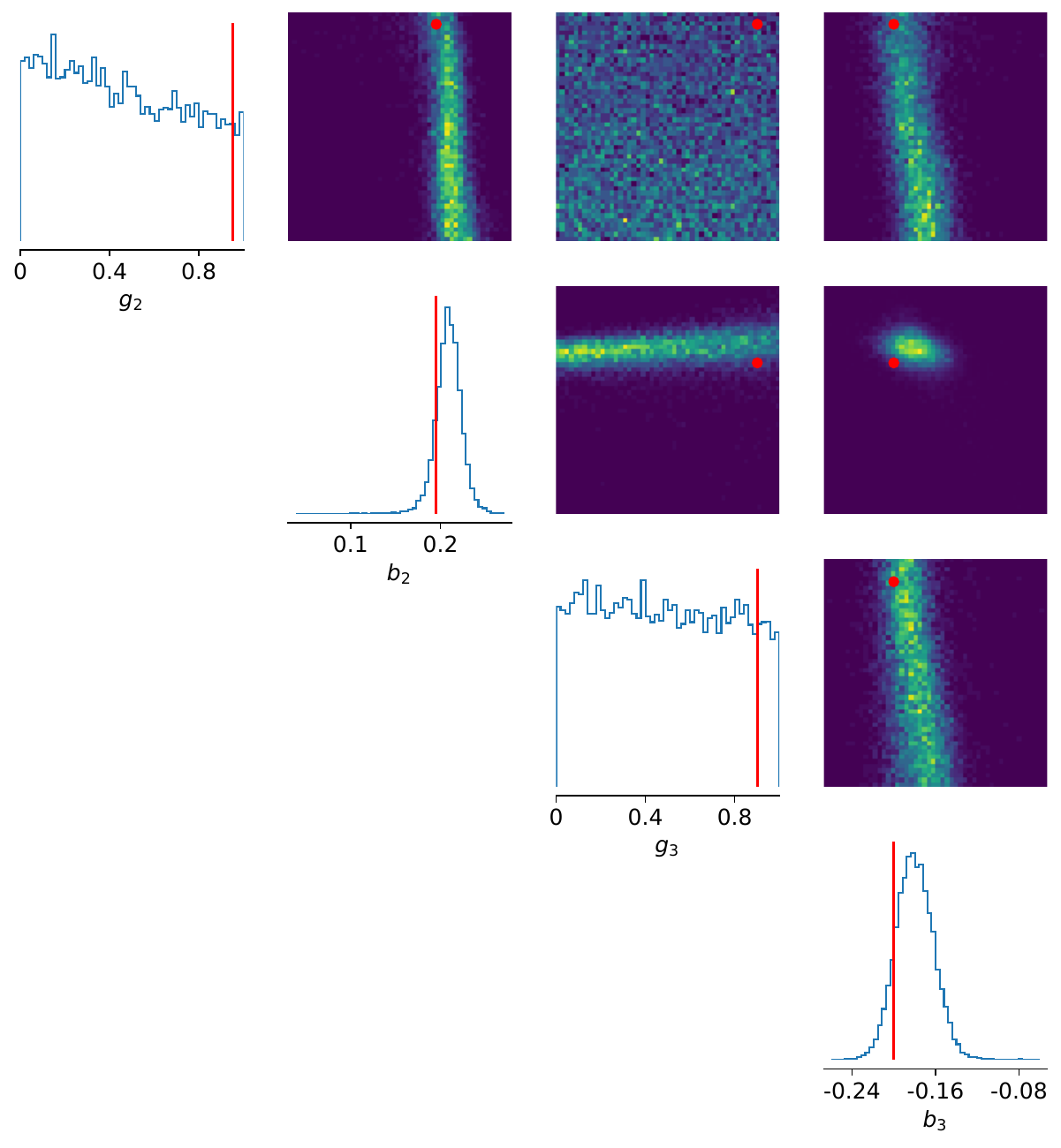}
\caption{Brock--Hommes, $\beta=60$.}
\label{fig:app_snpe_scratch_bh}
\end{subfigure}
\hfill
\begin{subfigure}[t]{0.48\textwidth}
\centering
\includegraphics[width=\linewidth]{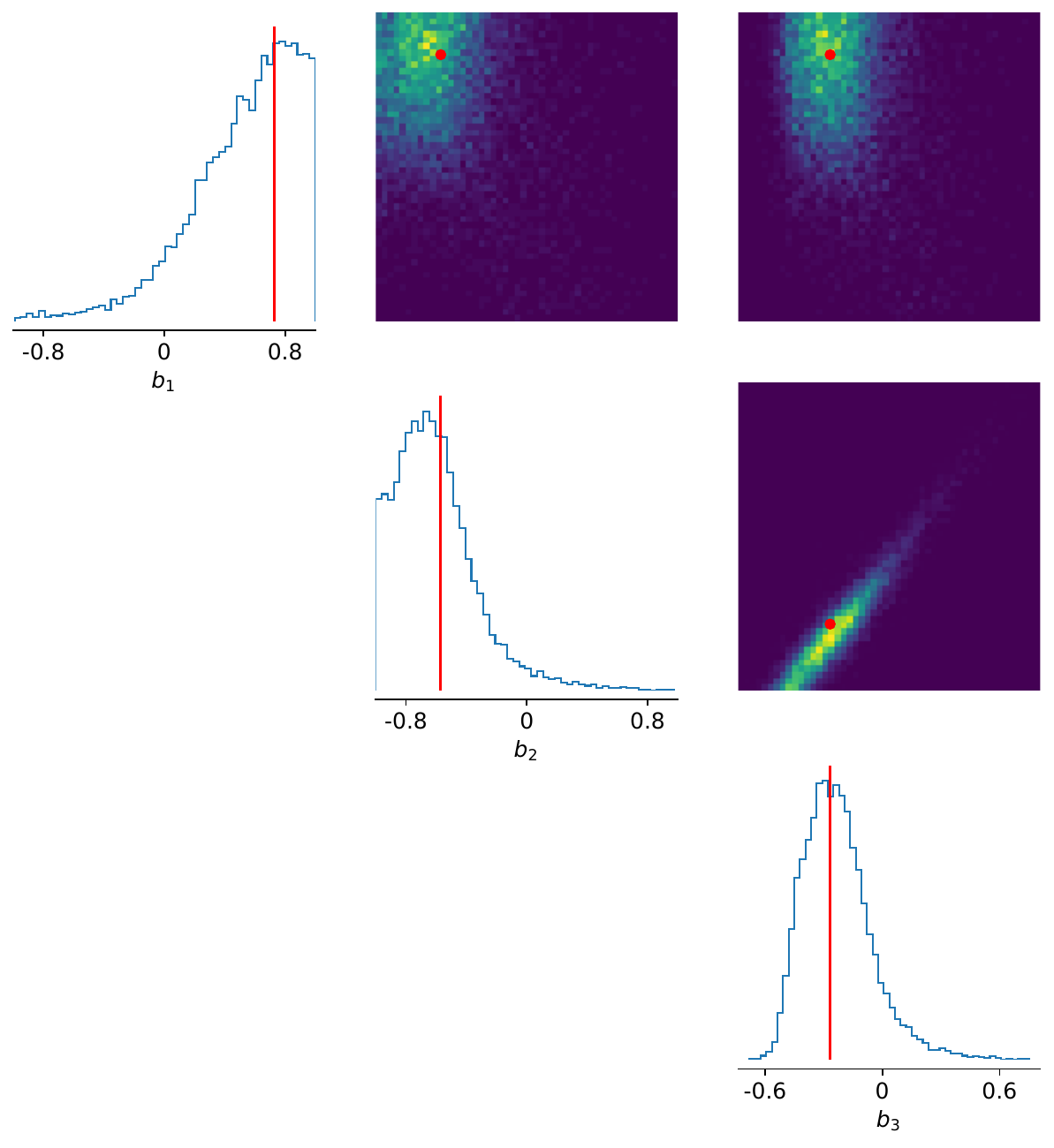}
\caption{MVGBM target observation.}
\label{fig:app_snpe_scratch_mvgbm}
\end{subfigure}
\caption{Posterior samples for SNPE from for the Brock-Hommes Model at $\beta=60$ (left) and target MVGBM (right) in Section~\ref{sec:experiments}, which use [500, 500, 500, 1000] as simulation budgets. Diagonal panels show marginal posterior distribution (blue curve), upper-triangular panels show bivariate joint density. Red lines and points denote the mean of the likelihood-based reference posterior. Darker (resp. lighter) colours indicate lower (resp. higher) posterior density in off-diagonal.}
\label{fig:app_snpe_scratch}
\end{figure}

Figure~\ref{fig:app_snpe_scratch} shows scratch SNPE fits for the two tractable target tasks. Each uses random initialisation and the same 2,500 target simulations as adaptation. Using 1,000 posterior and reference draws, WASS and MMD$^2$ are 0.7056 and 0.5761 for BH, and 0.2253 and 0.0581 for MVGBM, which follow the same calculations in Appendix~\ref{app:evaluation}. 

\section{Further experimental details and results}
\label{app:experiments}

\subsection{Brock-Hommes model}

For numerical experiments in Section~\ref{sec:experiments},the Brock-Hommes simulator uses $T=100$, $R=1.01$ and $\sigma=0.04$, with the first three price deviations initialised at zero. Source and target observations use $\beta=120$ and $\beta=60$, respectively, with the structural parameters and priors specified in Section~\ref{sec:bh-design}.

\subsection{Multivariate geometric Brownian motion}

In Section~\ref{sec:experiments}, the MVGBM simulator starts from $X_0=(10,10,10)$ and returns 100 three-dimensional states after removing the initial state. Its discrete log-state drift uses $c_i=\tfrac12\sum_j\sigma_{ji}^2$, a column-sum convention, in both simulation and likelihood evaluation. The source and target tasks use the same simulator, volatility matrix and prior.

\subsection{Franke-Westerhoff experimental details and results}
\label{app:fw_sbc}

The Franke-Westerhoff experiments are conducted without observation or parameter z-scoring. We multiply the returns by 100 and rescale the parameters to $u=(\alpha_w/15000,\eta,\sigma_c/5)\in[0,1]^3$. The simulator starts with zero prices, demands, and wealth, and produces 98 log returns when $T=100$. All methods use the same target simulations in the first round, while subsequent simulations depend on their respective fitted proposals. 

For evaluations in this experiment result, we use 100 independent target observations drawn from the prior predictive distribution. For all SNPE-TTT approaches training, we regenerate 999 independent posterior samples per observation using the available final checkpoints, without retraining. Further details of the sampling procedure are given in Appendix~\ref{app:sampling}, while the rank binning and reference bands are described in Section~\ref{sec:fw_metrics}. With this change, all 100 observations are included in the source SNPE SBC results, although the adaptation and scratch results still include only successfully completed cases.

\begin{figure}[!htbp]
\centering
\begin{subfigure}[t]{0.86\linewidth}
\centering
\includegraphics[width=\linewidth]{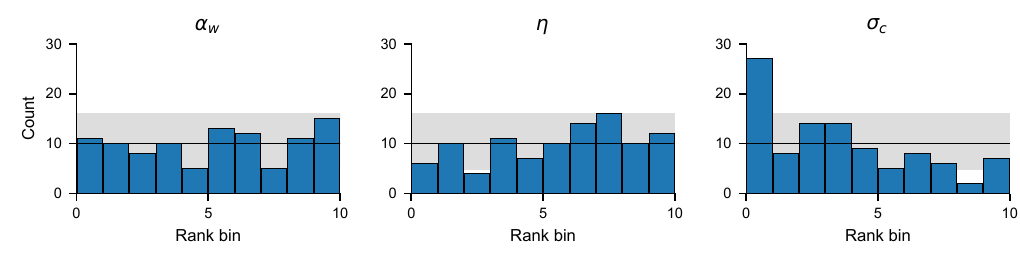}
\caption{SNPE for source FW model}
\end{subfigure}
\par\smallskip
\begin{subfigure}[t]{0.86\linewidth}
\centering
\includegraphics[width=\linewidth]{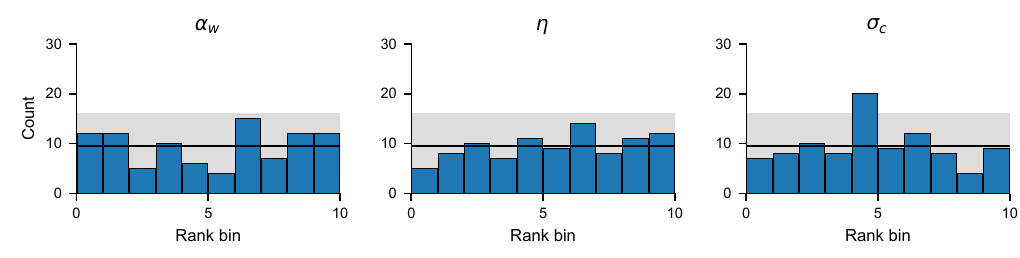}
\caption{SNPE for target FW model}
\end{subfigure}
\caption{Rank histograms generated according to the simulation-based calibration for source (top) and target (bottom) FW model. The shaded regions represent the symmetric 95\% point function of a Binomial random variable with success probability 1/P=1/10, that is, the bottom (resp. top) of the shaded regions is the inverse of the cumulative distribution function for a Binomial random variable at quantile 0.025 (resp. 0.975). This generally stands for the expected deviations from complete uniformity when the histograms are distributed uniformly.}
\label{fig:fw_sbc_baselines}
\end{figure}

Figure~\ref{fig:fw_sbc_baselines} also presents the SBC results for SNPE alongside the adaptation methods. Source-only applies the pretrained source network directly to each target observation, whereas scratch trains a new network from random initialization. The source SNPE results cover all 100 observations The source SNPE results therefore assess how well the pretrained network performs on the target observations, rather than its calibration in the original source setting.

Table~\ref{tab:fw_coverage} retains the central marginal intervals from all 5,000 saved draws and it is not recomputed from the new SBC samples. At approximately 100 observations, nominal 90\% coverage has a standard error near three percentage points.

\begin{table}[!htbp]
\centering
\scriptsize
\begin{tabular}{lr rrr rrr}
\toprule
& & \multicolumn{3}{c}{90\% coverage (\%)} & \multicolumn{3}{c}{Mean normalised width} \\
\cmidrule(lr){3-5}\cmidrule(lr){6-8}
Method & Cases & $\alpha_w$ & $\eta$ & $\sigma_c$ & $\alpha_w$ & $\eta$ & $\sigma_c$ \\
\midrule
Source-only & 77 & 87.0 & 92.2 & 83.1 & 0.875 & 0.742 & 0.284 \\
Scratch SNPE & 95 & 86.3 & 91.6 & 89.5 & 0.886 & 0.676 & 0.245 \\
SNPE-FPA & 98 & 85.7 & 92.9 & 88.8 & 0.871 & 0.615 & 0.225 \\
SNPE-LoRA & 99 & 86.9 & 88.9 & 92.9 & 0.866 & 0.614 & 0.236 \\
GradSubspace-FPA & 98 & 84.7 & 91.8 & 89.8 & 0.869 & 0.665 & 0.253 \\
GradSubspace-PEA & 98 & 86.7 & 89.8 & 91.8 & 0.868 & 0.673 & 0.255 \\
\bottomrule
\end{tabular}
\caption{FW marginal 90\% coverage and interval widths, normalised by the prior range, on completed observations.}
\label{tab:fw_coverage}
\end{table}

\subsection{Discrepancy calculations and repeated-fit results}
\label{app:evaluation}

Discrepancies use 1,000 draws per distribution on the original parameter scale. WASS uses optimal assignment with Euclidean costs. The biased MMD$^2$ estimator includes diagonal terms and uses the median approximate ground truth pairwise squared distance as its squared Gaussian bandwidth. Tests use untransformed values; logarithms are used only for plots. Each tractable task has 20 fits at one fixed target observation, measuring fitting variability rather than variability across tasks. Three two-sided Wilcoxon comparisons are performed within each model--metric family, using SciPy defaults and the Wilcox zero-difference convention. Holm correction covers those three comparisons at level 0.05. Table~\ref{tab:bh_complete_tests} reports all adjusted results and negative median differences favour the left method. 

\begin{table}[!htbp]
\centering
\small
\begin{tabular}{llrrr}
\toprule
& & \multicolumn{3}{c}{Contrast} \\
\cmidrule(lr){3-5}
Model & Metric & G-FPA $-$ FPA & PEA $-$ FPA & PEA $-$ G-FPA \\
\midrule
BH & WASS: Holm $p$ & 0.000629 & 0.002865 & 0.621513 \\
& Median difference & 0.041138 & 0.036054 & 0.001398 \\
& MMD$^2$: Holm $p$ & 1.000000 & 1.000000 & 1.000000 \\
& Median difference & $-0.007552$ & 0.015908 & $-0.009274$ \\
\midrule
MVGBM & WASS: Holm $p$ & 0.404900 & 0.398180 & 0.404900 \\
& Median difference & $-0.017316$ & $-0.030468$ & $-0.011050$ \\
& MMD$^2$: Holm $p$ & 0.116516 & 0.028307 & 0.595819 \\
& Median difference & $-0.011911$ & $-0.018546$ & $-0.003083$ \\
\bottomrule
\end{tabular}
\caption{Complete paired comparisons ($n=20$ each). FPA, G-FPA and PEA denote SNPE-FPA, GradSubspace-FPA and GradSubspace-PEA. Holm correction is separate for each model--metric family.}
\label{tab:bh_complete_tests}
\label{tab:mv_complete_tests}
\end{table}

\section{Posterior sampling}
\label{app:sampling}

\subsection{Sampling from the neural posterior}

For a fixed observation $y$, neural posterior samples are generated by drawing from the flow and rejecting values outside the prior support $\Theta$. The resulting density is
\begin{equation}
q_\phi^{\Theta}(\theta\mid y)
=\frac{q_\phi(\theta\mid y)\,\mathbf{1}_{\Theta}(\theta)}
{\int_{\Theta}q_\phi(\vartheta\mid y)\,d\vartheta}.
\end{equation}
The tractable repeated fits save 5,000 draws each, while the scratch SNPE in Appendix~\ref{app:snpe_posteriors} use 10,000. Completed FW fits save 5,000 independent rejection draws in the unit-cube parameterisation, which are mapped back to physical parameters. The FW interval summaries and recorded costs retain the archived rejection-sampling results. All six SBC histograms use the new 999-draw evaluations with the common rejection-sampling budget described in Appendix~\ref{app:fw_sbc}, but observations without final checkpoints remain excluded.

\subsection{Approximate ground-truth posteriors for likelihood-tractable models}
\label{app:references}

For the two tractable models, we have their avaiable likelihood so that MH targets $\pi(\theta\mid y)\propto p(y\mid\theta)p(\theta)$. Given the current state $\theta$ and a proposal $\theta'\sim r(\cdot\mid\theta)$, the acceptance probability is
\begin{equation}
a(\theta,\theta')=\min\!\left\{1,
\frac{\pi(\theta'\mid y)r(\theta\mid\theta')}
{\pi(\theta\mid y)r(\theta'\mid\theta)}\right\}.
\end{equation}
Proposals outside the prior support are rejected. This separates reference-posterior sampling from the direct neural posterior sampling described above.

BH uses Gaussian transition densities and one random-walk MH chain. The chain starts at $(0.9,0.2,0.9,-0.2)$. Its proposal covariance is the source-posterior covariance plus $10^{-6}I$, scaled by $2/\sqrt{4}$. With 100,000 iterations, 5,000 discarded draws and thinning by ten, it retains 9,500 samples.

MVGBM uses Gaussian log-state transitions with mean $Z_t+(b-c)\Delta t$ and covariance $\Delta t\,\sigma\sigma^\top$. Source and target chains start at their generating parameters, with Gaussian proposal standard deviation 0.05 and uniform support $[-1,1]^3$. Each uses 1,000 burn-in iterations and 5,000 unthinned draws; acceptance rates are 0.6832 and 0.6582. Neither benchmark retains effective sample sizes or multiple-chain diagnostics. The references therefore remain finite numerical benchmarks without convergence certification \citep{vehtari2021rank}.

\end{document}